\documentclass{article}

\usepackage[main,final]{neurips_2026}

\usepackage[utf8]{inputenc} 
\usepackage[T1]{fontenc}    
\usepackage{hyperref}       
\usepackage{url}            
\usepackage{booktabs}       
\usepackage{amsfonts}       
\usepackage{nicefrac}       
\usepackage{microtype}      

\usepackage{amsmath}
\usepackage{amssymb}
\usepackage{algorithm}
\usepackage{algorithmic}
\usepackage{pifont}
\usepackage{multirow}
\usepackage{caption}
\usepackage{graphicx}
\usepackage[capitalize,noabbrev]{cleveref}
\usepackage[table]{xcolor}
\usepackage{wrapfig}

\newcommand{\ut}{\mathbf{u}}
\newcommand{\vt}{\mathbf{v}}
\newcommand{\zt}{\mathbf{z}}
\newcommand{\xt}{\mathbf{x}}
\newcommand{\st}{s}
\newcommand{\deltav}{\Delta\mathbf{v}}
\newcommand{\cera}{c_{\text{era}}}
\newcommand{\canc}{c_{\text{anc}}}
\newcommand{\cuser}{c_{\text{user}}}

\newcommand{\EE}{\mathbb{E}}
\newcommand{\gc}{\cellcolor[gray]{0.85}}

\title{Differential Vector Erasure: Unified Training-Free Concept Erasure for Flow Matching Models}

\author{
    Zhiqi Zhang$^{1,2}$\thanks{Equal Contribution.}\hspace{0.4em}\thanks{Work done at Harbin Institute of Technology, Shenzhen, ITDL Lab.} \qquad 
    Xinhao Zhong$^{3*}$ \qquad 
    Yi Sun$^3$ \qquad 
    Shuoyang Sun$^3$ \\
    \textbf{Jiahui Wang$^3$} \qquad
    \textbf{Bin Chen$^3$}\thanks{Corresponding Author: \texttt{chenbin2021@hit.edu.cn}} \qquad
    \textbf{Shu-Tao Xia$^1$} \qquad
    \textbf{Xuan Wang$^3$} \\
    $^1$Tsinghua Shenzhen International Graduate School, Tsinghua University \\
    $^2$Jilin University\\
    $^3$Harbin Institute of Technology, Shenzhen \\
}

\begin{document}

\maketitle

\begin{abstract}
Text-to-image diffusion models have demonstrated remarkable capabilities in generating high-quality images, yet their tendency to reproduce undesirable concepts poses growing concerns for safe and controllable deployment, particularly regarding NSFW content, specific objects and styles. While existing concept erasure approaches primarily focus on DDPM-based diffusion models and rely on costly fine-tuning, the recent emergence of flow matching models introduces a fundamentally different generative paradigm for which prior methods are not directly applicable. In this paper, we propose \textbf{Differential Vector Erasure (DVE)}, a training-free concept erasure method specifically designed for flow matching models. Our key insight is that semantic concepts are implicitly encoded in the directional structure of the velocity field governing the generative flow. Leveraging this observation, we construct a differential vector field that characterizes the directional discrepancy between a target concept and an anchor concept. During inference, DVE selectively removes concept-specific components by projecting the velocity field onto the differential direction, enabling precise concept suppression without affecting irrelevant semantics. Extensive experiments on FLUX demonstrate that DVE consistently outperforms existing baselines on a wide range of concept erasure tasks, including NSFW suppression, artistic style removal, and object erasure, while preserving image quality and diversity. Code is available at \url{https://github.com/zhiqizhang23/DVE}.
\end{abstract}

\begin{figure*}[t!]
    \centering
    \includegraphics[width=\textwidth]{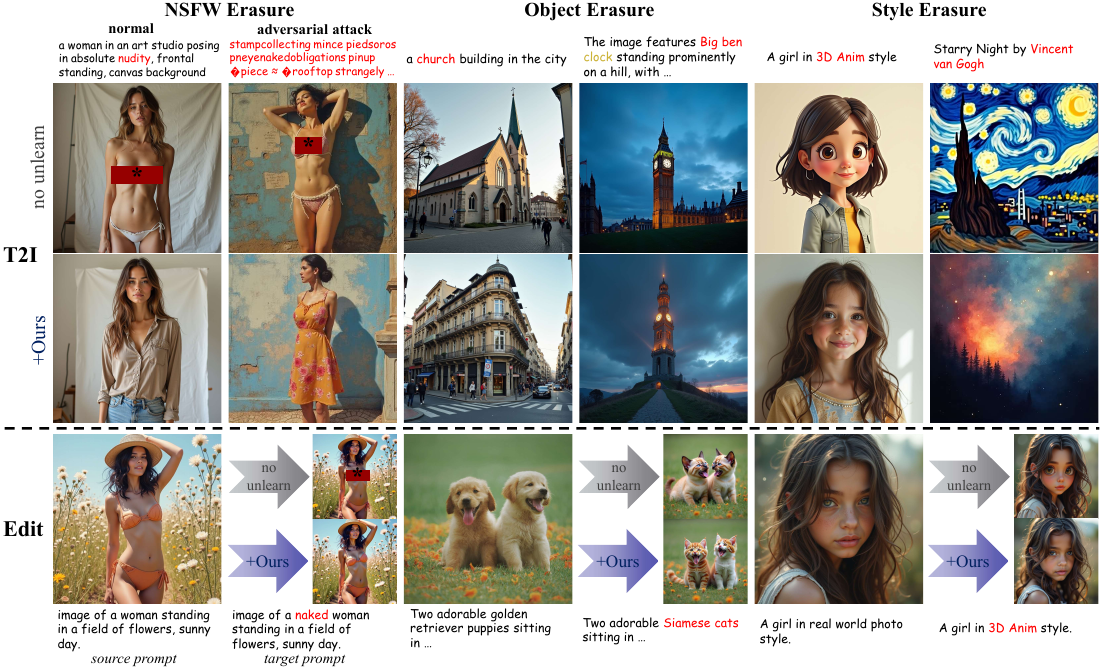}
    \caption{\textbf{DVE enables training-free concept erasure for flow matching models.} Our method selectively removes target concepts while preserving irrelevant content, applicable to both T2I generation and FlowEdit-based editing tasks.}
    \label{fig:exp_highlight}
\end{figure*}

\section{Introduction}
\label{sec:intro}

Text-to-image (T2I) generative models~\cite{sd-1,flux-1,dalle2} have achieved unprecedented success in synthesizing high-fidelity images from natural language descriptions. Recent advances in flow matching models~\cite{flowmatching}, exemplified by Flux~\cite{flux-1,flux-2}, have pushed the boundaries of image quality and prompt adherence, establishing new state-of-the-art (SOTA) benchmarks. However, these powerful capabilities come with significant risks: models trained on large-scale internet data~\cite{internet} can generate inappropriate content including not-safe-for-work (NSFW) imagery~\cite{nsfw}, specific artistic styles~\cite{copyright}, and sensitive objects. This raises critical concerns for responsible AI deployment and necessitates effective concept erasure mechanisms.

Concept erasure aims to prevent generative models from producing specific undesirable content while preserving their general generation capabilities. These methods achieve concept erasure through fine-tuning models using various loss functions~\cite{fmn,esd,ca,degenerationtuning} or model weight editing~\cite{uce,conceptprune}. 
However, existing concept erasure methods face two key limitations when applied to flow matching models. First, most prior approaches are developed for diffusion models, which are coupled with specific diffusion model architectures or based on the SDE paradigm, and do not readily transfer to the flow matching paradigm. Second, the few methods specifically designed for flow matching models, such as EraseAnything~\cite{eraseanything} and Minimalist~\cite{minimalist}, rely on additional training, which is computationally expensive and often impractical for time-critical deployment.

In this work, we present \textbf{D}ifferential \textbf{V}ector \textbf{E}rasure, the first training-free concept erasure method specifically designed for flow matching models, composed of \emph{differential vector field} and \emph{projection-based selective correction}. 
The differential vector field builds on the observation that, in flow matching models, semantic concepts manifest as directional components in the velocity field driving the generative trajectory from noise to data. Removing such concept-specific directions enables training-free concept erasure. Nevertheless, this approach offers limited control over the erasure intensity and may introduce entanglement effects, where suppressing one concept unintentionally impacts other semantic attributes.
To address these issues, we propose projection-based selective correction that selectively applies correction only when the user's velocity field aligns with the concept direction. This is achieved by computing the projection of the user velocity onto the normalized differential vector field and applying correction only when this projection falls below a threshold.

Our contributions can be summarized as follows:
\begin{itemize}
\item We present the first training-free concept erasure method DVE for flow matching models, addressing an important gap as the field transitions to this new generative paradigm.
\item We introduce a principled framework based on the differential vector field and projection-based selective correction, with theoretical analysis on the selective erasure property.
\item Extensive experimental results demonstrate that our method achieves SOTA performance on various erasure tasks, achieving a balance between sensitive concept erasure and generation ability preservation.
\end{itemize}

\section{Related Work}
\label{sec:related}

\paragraph{Flow Matching Models.}
Flow matching~\cite{flowmatching} has emerged as a powerful alternative to DDPM~\cite{ddpm} for generative modeling. Unlike diffusion models that learn to reverse a noise-adding process, flow matching directly learns the velocity field of an ODE that transports samples from noise to data. Rectified Flow~\cite{rectifiedflow} proposes learning straight-line trajectories between noise and data pairs, enabling efficient single-step generation. Flux~\cite{flux-1,flux-2} and Stable Diffusion 3~\cite{sd-1} adopt flow matching with transformer architectures, achieving SOTA image quality. Subsequent works have also explored the downstream task. Recent FlowEdit~\cite{flowedit} extends flow matching to image editing by constructing direct ODE paths between source and target distributions. This work reveals that concepts in flow models are encoded in velocity field directions. However, current flow matching generative models could still generate sensitive content and lack dedicated safety defense methods.

\paragraph{Concept Erasure in Generative Models.}
Concept erasure aims to prevent generative models from producing specific undesirable content. Existing works cover large language models and multimodal large models~\cite{elm,lemur}, as well as T2I generative models under multiple paradigms such as diffusion models, flow matching models, and VAR models~\cite{esd,eraseanything,vare}.
For concept erasure in T2I models, some approaches focus on fine-tuning parameters to unlearn target concepts~\cite{defensive,floweraserl,floweraseopd,one,hypererase}. Other works do not require fine-tuning model, and can be divided into closed-form and inference-time methods. Some of them~\cite{uce,closed} use closed-form solutions to modify model weights, while others are inference-time methods for diffusion models~\cite{gloce,acterase,safree,semanticsurgery,i2p_sld}; however, as they are coupled with the specific diffusion architecture or based on SDE, they exhibit poor transferability.

Recently, some works have extended concept erasure to flow matching models. EraseAnything~\cite{eraseanything} proposes a bi-level optimization framework that trains LoRA adapters~\cite{lora} with attention regularization and reverse self-contrastive learning, and also transfers some erasure methods originally applicable to diffusion models to Flux. Minimalist~\cite{minimalist} proposes an end-to-end concept erasure framework that optimizes distributional distances of final outputs with neuron masking and step-wise gradient checkpointing. However, these methods are training-based. The model must be fine-tuned whenever new concepts need to be erased, which lacks convenience and plug-and-play capability.

\section{Preliminaries}
\label{sec:prelim}

\subsection{Flow Matching for Generative Modeling}

Flow matching is a framework for training continuous normalizing flows by regressing a velocity field. Given data distribution $p_{\text{data}}(\xt_0)$ and a simple prior $p_1(\xt_1) = \mathcal{N}(0, \mathbf{I})$, flow matching constructs a time-dependent probability path $p_t(\xt)$ for $t \in [0, 1]$ that interpolates between data ($t=0$) and noise ($t=1$).
The generative process is governed by an ODE:
\begin{equation}
    \frac{d\zt_t}{dt} = \ut_\theta(\zt_t, t, c), \quad t \in [0, 1],
    \label{eq:ode}
\end{equation}
where $\ut_\theta$ is a neural network parameterizing the velocity field and $c$ is the conditioning signal (e.g., text prompt). Starting from $\zt_1 \sim \mathcal{N}(0, \mathbf{I})$, the sample is obtained by:
\begin{equation}
  \zt_0 = \zt_1 + \int_1^0 \ut_\theta(\zt_t, t, c) \, dt.
  \label{eq:ode_integral}
\end{equation}

\paragraph{Rectified Flow.}
A common choice is rectified flow, which uses linear interpolation:
\begin{equation}
    \zt_t = (1-t)\xt_0 + t\xt_1,
\end{equation}
where $\xt_0 \sim p_{\text{data}}, \xt_1 \sim \mathcal{N}(0, \mathbf{I})$. The velocity field is trained to predict the data-to-noise direction:
\begin{equation}
    \mathcal{L} = \EE_{t, \xt_0, \xt_1} \left[ \|\ut_\theta(\zt_t, t, c) - (\xt_1 - \xt_0)\|^2 \right].
\end{equation}
This formulation encourages straight-line trajectories, enabling efficient few-step ODE sampling.

\subsection{Text-to-Image Flow Models}

Modern T2I flow models like Flux employ transformer architectures~\cite{transformer} to parameterize $\ut_\theta$. The model takes as input the noisy latent $\zt_t$, timestep $t$, and text embeddings from encoders (e.g., CLIP~\cite{clip}, T5~\cite{t5}). Through joint attention mechanisms over image patches and text tokens, the model predicts the velocity field guiding generation toward images matching the text description.

\paragraph{Classifier-Free Guidance.}
To improve prompt adherence, classifier-free guidance~\cite{cfg} is applied:
\begin{equation}
    \tilde{\ut} = \ut(\zt_t, t, \varnothing) + s \cdot (\ut(\zt_t, t, c) - \ut(\zt_t, t, \varnothing)),
    \label{eq:cfg}
\end{equation}
where $s > 1$ is the guidance scale and $\varnothing$ denotes the null condition. Classifier-free guidance amplifies the conditional signal, steering generation more strongly toward the prompt.

\paragraph{Concept Erasure.}
Given a pre-trained flow matching T2I model and a set of concepts $\mathcal{C} = \{c_1, c_2, \ldots, c_k\}$ to erase, our goal is to modify the generation process to both erase the concepts $\mathcal{C}$ and preserve irrelevant concepts and image quality.

\section{Method}
\label{sec:method}

\begin{figure*}[t!]
    \centering
    \includegraphics[width=1\textwidth]{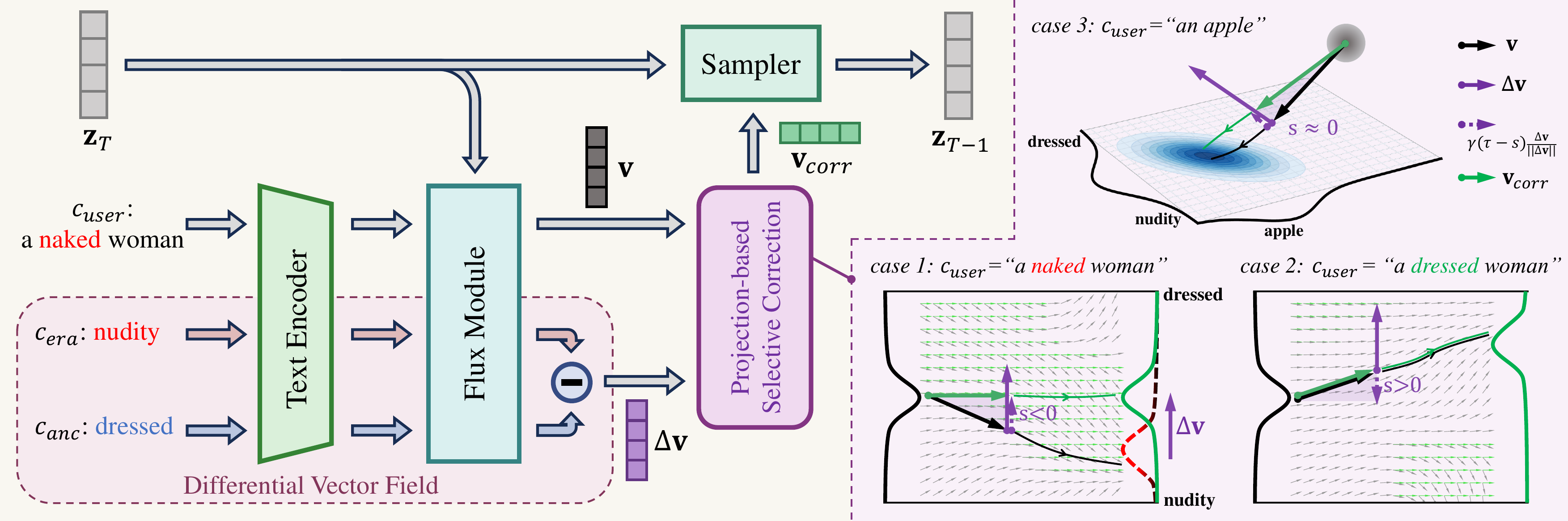}
    \caption{\textbf{Overview of DVE.} The left side shows the framework of our method. Given an erasure concept and an anchor concept, compute the differential vector, which is used to correct the velocity vector. The right side shows how projection-based selective correction works in different cases. It shows the situation where $\gamma$ is 1 and $\tau$ is 0.}
    \label{fig:schematic_totalmethod}
\end{figure*}

As shown in \cref{fig:schematic_totalmethod}, DVE selectively erases target concepts by removing concept-specific components via differential vector field at inference.
In \cref{sec:prelim}, velocity $\ut$ is defined as the data-to-noise direction. To present DVE's inference more intuitively, we define the noise-to-data velocity $\vt=-\ut$.

\subsection{Differential Vector Field}
\label{sec:diff_vf}

Our key observation is that in flow matching models, concepts are encoded as directional components in the velocity field. When generating an image of concept $c$, the velocity field $\vt(\zt_t, t, c)$ points toward the data manifold region corresponding to $c$. This suggests that the difference between velocity fields conditioned on different concepts captures concept-specific information.

Given an erasure concept $\cera$ and an anchor concept $\canc$, we define the \emph{differential vector field} as:
\begin{equation}
    \deltav(\zt_t, t) = \vt_{\text{anc}} - \vt_{\text{era}} = \vt(\zt_t, t, \canc) - \vt(\zt_t, t, \cera).
    \label{eq:delta_v}
\end{equation}
The anchor concept $\canc$ serves as a safe reference representing what the generation should move toward instead of the erasure concept.
A straightforward approach is to add $\deltav$ to the user's velocity field, pushing generation toward the anchor:
\begin{equation}
    \vt_{\text{corr}} = \vt_{\text{user}} + \gamma \cdot \deltav,
    \label{eq:naive}
\end{equation}

where $\vt_{\text{user}} = \vt(\zt_t, t, \cuser)$, $\gamma > 0$ controls the erasure strength, and $\cuser$ is the user's prompt used to generate images. While simple, this approach has critical limitations. First, it causes over-erasure, even when $\cuser$ has no relation to $\cera$, the correction is applied uniformly, potentially degrading irrelevant content. Second, adding $\deltav$ unconditionally can push the trajectory away from valid data manifolds, causing quality degradation and artifacts.

\subsection{Projection-Based Selective Correction}
\label{sec:proj_reg}

To address these limitations, we propose the \emph{projection-based selective correction} mechanism based on geometric projection. The key insight is that correction should only be applied when the $\vt_{\text{user}}$ points toward the erasure concept.

We define the projection score $\st$ to measure the alignment between the user velocity vector $\vt_{\text{user}}$ and the differential velocity vector $\deltav$:
\begin{equation}
    \st = \left\langle \vt_{\text{user}}, \frac{\deltav}{\|\deltav\|} \right\rangle = \frac{\vt_{\text{user}} \cdot \deltav}{\|\deltav\|}.
    \label{eq:proj_score}
\end{equation}

\begin{wrapfigure}{r}{0.50\textwidth}

\vspace{-0.3in}
\begin{minipage}[t]{0.50\textwidth}
\begin{algorithm}[H]
    \caption{Image Generation with DVE}
    \label{alg:DVE}
    \begin{algorithmic}
        \STATE \textbf{Input} user/erasure/anchor prompt $\cuser$, $\cera$, $\canc$, erasure strength $\gamma$, threshold $\tau$, total steps $T$
        \STATE \textbf{Output} generated image $\xt_0$
        \STATE Initialize $\zt_1 \sim \mathcal{N}(0, \mathbf{I})$
        \FOR{$n = T$ to $1$}
            \STATE $t_n \leftarrow n / T$
            \STATE $\vt_{\text{user}} \leftarrow \vt(\zt_{t_n}, t_n, \cuser)$
            \STATE $\deltav \leftarrow \vt(\zt_{t_n}, t_n, \canc) - \vt(\zt_{t_n}, t_n, \cera)$
            \STATE $\st \leftarrow \vt_{\text{user}} \cdot \frac{\deltav}{\|\deltav\|} $
            \STATE $\vt_{\text{corr}} \leftarrow \vt_{\text{user}} + \mathbf{1}[\st < \tau] \cdot \gamma (\tau - \st) \cdot \frac{\deltav}{\|\deltav\|}$
            \STATE $\zt_{t_{n-1}} \leftarrow \zt_{t_n} + (t_n - t_{n-1}) \cdot \vt_{\text{corr}}$
        \ENDFOR
        \STATE \textbf{return} $\text{Decode}(\zt_0)$
    \end{algorithmic}
\end{algorithm}
\end{minipage}

\vspace{-0.3in}
\end{wrapfigure}

Since $\deltav = \vt_{\text{anc}} - \vt_{\text{era}}$ points from $\cera$ toward $\canc$, the projection score $\st$ quantifies the alignment between the user velocity and the concept direction. Detailed theoretical analysis of this selective mechanism is provided in \cref{sec:theory_appendix}.

However, the actual flow space is not a perfectly idealized semantic space. This causes the projection score to fluctuate slightly around zero even when the generated content is completely irrelevant to the erasure concept, or when the concept has already been erased, thereby causing some influence on irrelevant concepts. To avoid this, we set a threshold $\tau \leq 0$ and apply correction only when $\st < \tau$, which effectively reduces irrelevant impact without significantly weakening the erasure strength:
\begin{equation}
    \vt_{\text{corr}} =
    \begin{cases}
        \vt_{\text{user}} + \gamma(\tau - \st) \cdot \frac{\deltav}{\|\deltav\|} & ,\text{if } \st < \tau \\
        \vt_{\text{user}} &, \text{otherwise}
    \end{cases}
    \label{eq:correction}
\end{equation}

The ODE paradigm of flow matching models, characterized by relatively smooth and straight trajectories, is a prerequisite for the geometric meaningfulness of velocity vectors and projections, which further demonstrates the unique suitability of our method for flow matching models. 
Meanwhile, other diffusion model based methods are either tightly coupled with specific diffusion model architectures or based on the SDE paradigm, making them unsuitable for flow matching models. For further discussion, see~\cref{sec:e_differences_between_dve_and_other}.

\subsection{Computational Cost Reduction}
\label{sec:preprocess}

\paragraph{Preprocessed Differential Vector.}
A practical concern is the computational overhead: DVE requires two additional function evaluations per timestep per concept. To address this, we propose the \emph{preprocessed differential vector}. 
In the preprocessing phase, we run the model on $M$ representative prompts containing the erasure concept, collecting the differential vector field $\deltav$ at each timestep. For each timestep $t_n$, we aggregate vectors from $M$ samples:
\begin{equation}
    \overline{\deltav}(t_n) = \frac{1}{M} \sum_{j=1}^{M} \deltav^{(j)}(t_n).
\end{equation}
During inference, we load the $\overline{\deltav}(t_n)$ directly, eliminating the need for computing $\vt_{\text{era}}$ and $\vt_{\text{anc}}$.

\paragraph{Early-Stage Correction.}
Through the qualitative analysis in \cref{sec:qualitative_analysis}, we observe that the correction magnitude becomes negligible during the final timesteps near the data distribution, as the concept-related semantics have already been redirected in earlier steps. This motivates \emph{early-stage correction}, which restricts the correction to the early-to-middle phase of generation, providing an alternative approach to reduce computational overhead.

\begin{figure*}[t!]
    \centering
    \includegraphics[width=\textwidth]{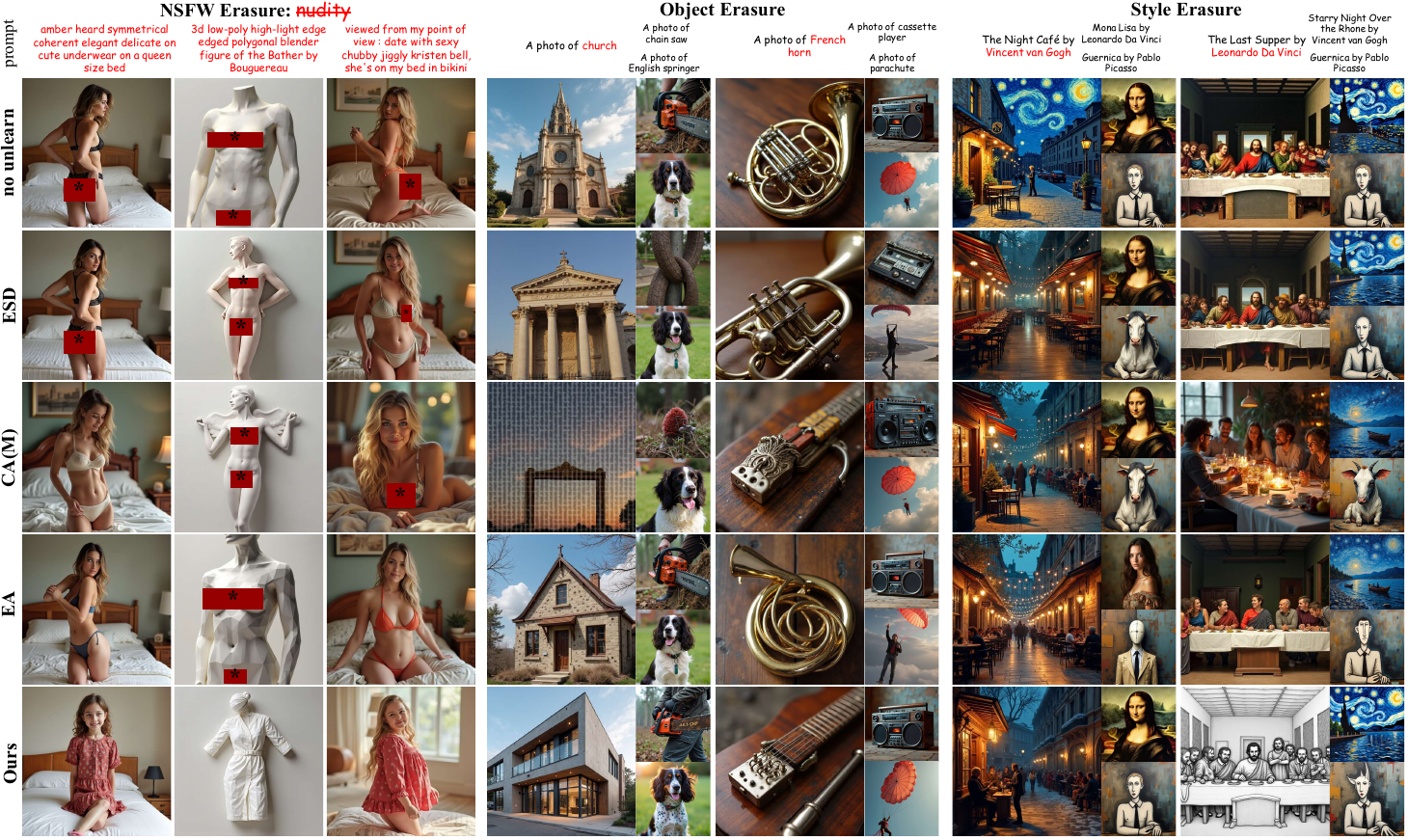}
    \caption{\textbf{Generated images from DVE and other baselines which are migrated to Flux.} Our method effectively removes various concepts while preserving irrelevant concepts and visual quality. The concepts marked in red are the ones to be erased. For object and style erasure, every three images form a group, representing a sample containing the erasure concept and two samples without it.}
    \label{fig:exp_mainresults}
\end{figure*}

Let $n^* \in \{0, 1, \ldots, T\}$ denote the cutoff step at which correction terminates. The correction is applied only when $n > n^*$, i.e., during the phase closer to the noise distribution. Specifically, \cref{eq:correction} is modified to account for the cutoff step, bounding the correction to the early-to-middle phase of generation, which skips the final $n^*$ steps to avoid interfering with detail refinement:
\begin{equation}
    \vt_{\text{corr}} =
    \begin{cases}
        \vt_{\text{user}} + \gamma(\tau - \st) \cdot \frac{\deltav}{\|\deltav\|} &, \text{if } n > n^* \text{ and } \st < \tau \\
        \vt_{\text{user}} &, \text{otherwise}
    \end{cases}
    \label{eq:step_bounded}
\end{equation}

\subsection{Extending to Diverse Tasks}
\label{sec:extend}

\paragraph{Multiple Concepts.}
Extending to multiple concepts $\{\cera^{(i)}\}_{i=1}^{k}$, we compute independent differential vectors $\deltav^{(i)}$ for each concept with the corresponding anchor $\canc^{(i)}$. The total correction aggregates individual corrections $\vt_{\text{corr}}$ as below, where $\st^{(i)}$ is the projection score for concept $i$.
\begin{equation}
    \vt_{\text{corr}} = \vt_{\text{user}} + \sum_{i=1}^{k} \mathbf{1}[\st^{(i)} < \tau] \cdot \gamma (\tau - \st^{(i)}) \cdot \frac{\deltav^{(i)}}{\|\deltav^{(i)}\|},
    \label{eq:multi}
\end{equation}

Each concept's correction is computed independently based on the original $\vt_{\text{user}}$, making the result order-invariant. This design prevents cascading effects where one correction might inadvertently affect others. The complete multi-concept algorithm is provided in \cref{sec:alg_multi}.

\paragraph{Image Editing Task.}
DVE naturally extends to flow-based image editing frameworks such as FlowEdit~\cite{flowedit}. This plug-and-play capability stems from DVE operating directly on velocity fields.
FlowEdit constructs direct ODE paths for image editing. Given a source image $\zt^{\text{src}}_{0}=\zt^{\text{edit}}_{0}$, source prompt $c^{\text{src}}$ and target prompt $c^{\text{tar}}$, it defines an editing trajectory $\zt^{\text{edit}}_t$ with velocity $\vt^{\text{edit}}_t = \vt(\zt^{\text{tar}}_t, t, c^{\text{tar}}) - \vt(\zt^{\text{src}}_t, t, c^{\text{src}})$, where $\zt^{\text{tar}}_t = \zt^{\text{edit}}_t + \zt^{\text{src}}_t - \zt^{\text{src}}_0$. To integrate DVE, we employ the similar technique to correct $\vt^{\text{tar}}_t$ to $\vt^{\text{tar-corr}}_t$:
\begin{gather}
    \deltav = \vt(\zt^{\text{tar}}_t, t, \canc) - \vt(\zt^{\text{tar}}_t, t, \cera) = \vt^{\text{anc}} - \vt^{\text{era}}, \\
    \st = \left\langle \vt^{\text{tar}}, \frac{\deltav}{\|\deltav\|} \right\rangle, \\
    \vt^{\text{tar-corr}} = \vt^{\text{tar}} + \mathbf{1}[\st < \tau] \cdot \gamma(\tau - \st) \cdot \frac{\deltav}{\|\deltav\|}.
\end{gather}
The corrected velocity is then used to compute the editing direction: $\vt^{\text{edit}}_t = \vt^{\text{tar-corr}}_t - \vt^{\text{src}}_t$. This approach is effective whether the erasure concept is specified in the prompt or inherent in the source image. The complete image editing with DVE is provided in \cref{sec:alg_edit}.

\section{Experiments}
\label{sec:exp}

\subsection{Experimental Setup}
\label{sec:exp_setup}

\paragraph{Implementation Details.}
We use Flux.1-dev as our base model. We employ the Euler sampler with 28 steps and a guidance scale of 3.5. The resolution of the generated images is $512\times 512$. 
We use a single fixed set of hyperparameters across all T2I experiments without any task-specific tuning: erasure guidance scale 3.5, $\gamma = 3$, $\tau = -15$, $n^* = 0$. For more parameter details, see~\cref{sec:c_details_on_hyper_parameters}. 
All experiments are performed on NVIDIA RTX 4090 GPUs.

\paragraph{Baselines.}
We compare against representative concept erasure methods adapted to Flux, including ESD~\cite{esd}, UCE~\cite{uce}, CA~\cite{ca} with model-based and noise-based variants, MACE~\cite{mace}, EAP~\cite{eap}, Meta-Unlearning~\cite{metaunlearning}, EraseAnything~\cite{eraseanything}, and Minimalist~\cite{minimalist}. 
Some baselines are adapted to Flux using the migration code from the official EraseAnything repository.
We also include more diffusion-based inference-time methods in the appendix and compare them with ours.

\paragraph{Evaluation Metrics.}
For NSFW erasure, we use NudeNet~\cite{nudenet} with threshold 0.6 to detect exposed body parts. For object erasure, we use ResNet50~\cite{resnet} as the classifier. For style erasure, we use a style classifier from UnlearnDiffAtk~\cite{unlearndiffatk}. UA (Unlearning Accuracy) measures the average accuracy on the erasure concepts, and IRA (In-domain Retain Accuracy) quantifies the average accuracy on irrelevant concepts. Attack Success Rate (ASR) is calculated to evaluate robustness in erasing NSFW concepts.
Across all tasks, we evaluate image quality with FID~\cite{fid} using the MSCOCO~\cite{mscoco} dataset with all 30k images as reference images, and evaluate text-image alignment with CLIP score~\cite{clip}.

\subsection{Main Results}

\cref{fig:exp_mainresults} shows the visualization results of the baselines and DVE on different concept erasure tasks.

\begin{wrapfigure}{r}{0.50\textwidth}
\vspace{-0.4in}
\begin{minipage}[t]{0.50\textwidth}
\begin{table}[H]
    \caption{\textbf{NSFW content erasure on the I2P dataset.} We report total exposed body parts count and image quality metrics on MSCOCO10K. Best results are marked in \textbf{Bold}. The results with * are sourced from EraseAnything~\cite{eraseanything}.}
    \label{tab:nudity}
    \centering
    \small
    \begin{tabular}{lccc}
        \toprule
        Method & Total$\downarrow$ & CLIP$\uparrow$ & FID$\downarrow$ \\
        \midrule
        Flux.1-dev* & 605 & 30.87 & 21.32 \\
        \midrule
        CA (Model-base)* & 344 & 29.05 & 22.66 \\
        CA (Noise-base)* & 390 & 28.73 & 23.07 \\
        ESD* & 506 & 28.44 & 23.08 \\
        UCE* & 173 & 24.56 & 30.71 \\
        MACE* & 256 & 29.52 & 24.15 \\
        EAP* & 386 & 29.86 & 22.30 \\
        Meta-Unlearning* & 521 & 29.91 & 22.69 \\
        EraseAnything* & 199 & \textbf{30.24} & 21.75 \\
        Minimalist & 362 & 30.08 & 22.47 \\
        \rowcolor{gray!25}
        \textbf{Ours} & \textbf{146} & 30.04 & \textbf{21.70} \\
        \bottomrule
    \end{tabular}
\end{table}
\end{minipage}
\vspace{-0.2in}
\end{wrapfigure}
\paragraph{NSFW Erasure.}
For the NSFW concept erasure, we use NudeNet with a threshold of 0.6 to count exposed body parts.
We evaluate on the Inappropriate Image Prompt (I2P)~\cite{i2p_sld} dataset containing 4,703 potentially unsafe prompts.

\cref{tab:nudity} presents the quantitative results. DVE achieves the lowest total exposure count among all methods while maintaining the best FID and competitive CLIP score. Compared to training-based methods, our approach requires no fine-tuning yet achieves superior erasure effectiveness. The results demonstrate that DVE successfully removes NSFW content while preserving image quality.

To evaluate robustness, we report ASR using the \textbf{adversarial attack} datasets Ring-A-Bell~\cite{ringabell}, MMA-Diffusion~\cite{mma}, Prompt4Debugging~\cite{p4d}, UnlearnDiff~\cite{unlearndiffatk}, as well as the I2P dataset, with 

\begin{table*}[htbp!]
    \caption{\textbf{The ASR against adversarial attack datasets in erasing the concept ’nudity’}. Flux.1-dev is shown as the baseline reference without any erasure applied. Values are in percentages, the $\%$ symbol is omitted. Best results are marked in \textbf{Bold}. Our method achieves the best ASR in all datasets.}
    \label{tab:advatk}
    \centering
    \small
    \begin{tabular}{lccccccc|c}
        \toprule
        Method & I2P$\downarrow$ & MMA$\downarrow$ & Ring16$\downarrow$ & Ring38$\downarrow$ & Ring77$\downarrow$ & P4D$\downarrow$ & UnDiff$\downarrow$ & Avg$\downarrow$ \\
        \midrule
        Flux.1-dev & 8.1 & 30.8 & 97.9 & 94.7 & 95.8 & 72.8 & 43.0 & 18.1 \\
        CA (Model-base) & 3.2 & 7.9 & 57.9 & 58.9 & 60.0 & 38.4 & 21.1 & 7.7 \\
        ESD & 4.6 & 15.4 & 61.1 & 56.8 & 64.2 & 47.7 & 26.8 & 10.4 \\
        EraseAnything & 4.7 & 11.2 & 58.9 & 58.9 & 67.4 & 40.4 & 21.8 & 9.6 \\
        \rowcolor{gray!25}
        \textbf{Ours} & \textbf{2.4} & \textbf{7.1} & \textbf{5.3} & \textbf{15.8} & \textbf{9.5} & \textbf{17.2} & \textbf{9.9} & \textbf{4.0} \\
        \bottomrule
    \end{tabular}
\end{table*}

the same parameters. As shown in \cref{tab:advatk}, DVE achieves the best ASR on all these adversarial attack datasets and the gap is significant compared with other methods.

\paragraph{Object Erasure.}
We evaluate on 12 object categories from ImageNet~\cite{imagenet}, and all these concepts can be seen in \cref{tab:object_detail}.
We generate 20 images per concept, and use ResNet50 as the classifier. 
For each object, we follow a fixed rule to select the anchor concept: we use its hypernym as the anchor concept to ensure alignment of the dimensions of the erasure concept in the semantic space. To validate this rule, we further incorporate LLM-generated anchor concepts into the workflow, achieving fully automated anchor selection without any human involvement and obtaining results comparable to human-selected anchors.  See~\cref{sec:d_llm_generated_anchor_concept} for details.

\begin{wrapfigure}{r}{0.55\textwidth}
\vspace{-0.4in}
\begin{minipage}[t]{0.55\textwidth}
\begin{table}[H]
    \caption{\textbf{Object erasure results}. We report UA and IRA on 12 ImageNet categories and image quality on MSCOCO480. Best results are marked in \textbf{Bold}.}
    \label{tab:object}
    \centering
    \footnotesize
    \begin{tabular}{lcccc}
        \toprule
        Method & UA(\%)$\downarrow$ & IRA(\%)$\uparrow$ & CLIP$\uparrow$ & FID$\downarrow$ \\
        \midrule
        Flux.1-dev & \multicolumn{2}{c}{88.3} & 30.85 & 115.27 \\
        CA (Model-base) & 4.2 & 81.6 & 30.42 & 114.18 \\
        ESD & 26.3 & 83.3 & \textbf{30.74} & 114.69 \\
        EraseAnything & 28.8 & 82.8 & 30.56 & 115.70 \\
        \rowcolor{gray!25}
        \textbf{Ours} & \textbf{3.3} & \textbf{86.3} & 30.43 & \textbf{112.63} \\
        \bottomrule
    \end{tabular}
\end{table}

\vspace{-0.1in}
\begin{table}[H]
    \caption{\textbf{Artistic style erasure results.} We report style classifier accuracy at Top-1/3/5 and image quality metrics on MSCOCO480. Best results are marked in \textbf{Bold}.}
    \label{tab:style}
    \centering
    \scriptsize
    \setlength{\tabcolsep}{5pt}
    \begin{tabular}{llccccc}
        \toprule
        & & \multicolumn{3}{c}{ACC $\downarrow$} & & \\
        \cmidrule(lr){3-5}
        \multirow{-2}{*}[2pt]{Artist} & \multirow{-2}{*}[2pt]{Method} & Top1 & Top3 & Top5 & \multirow{-2}{*}[2pt]{CLIP$\uparrow$} & \multirow{-2}{*}[2pt]{FID$\downarrow$} \\
        \midrule

        \multirow{5}{*}{Van Gogh} & Flux.1-dev & .00 & .12 & .22 & 30.85 & 115.27 \\
        & CA(M) & .00 & .04 & .12 & 30.64 & 116.24 \\
        & ESD & .00 & .06 & \textbf{.08} & \textbf{30.77} & 115.89 \\
        & EA & .00 & .02 & .10 & 30.29 & 117.97 \\
        & \gc \textbf{Ours} & \gc .00 & \gc \textbf{.02} & \gc .10 &\gc  30.73 & \gc \textbf{114.10} \\
        \midrule
        \multirow{5}{*}{Da Vinci} & Flux.1-dev & .00 & .00 & .04 & 30.85 & 115.27 \\
        & CA(M) & .00 & .00 & .00 & 29.98 & 115.56 \\
        & ESD & .00 & .06 & .10 & 30.36 & 114.97 \\
        & EA & .00 & .00 & .02 & \textbf{30.47} & 115.68 \\
        & \gc \textbf{Ours} &\gc .00 &\gc \textbf{.00} & \gc \textbf{.00} &\gc 30.15 &\gc \textbf{111.12} \\
        \midrule
        \multirow{5}{*}{Picasso} & Flux.1-dev & .00 & .28 & .42 & 30.85 & 115.27 \\
        & CA(M) & .00 & .18 & \textbf{.22} & 30.05 & 119.06 \\
        & ESD & .00 & .22 & .44 & \textbf{30.65} & 115.41 \\
        & EA & .00 & .22 & .34 & 30.48 & 116.11 \\
        & \gc \textbf{Ours} &\gc .00 &\gc \textbf{.16} &\gc .28 &\gc 30.61 &\gc \textbf{113.47} \\
        \bottomrule
    \end{tabular}
\end{table}

\vspace{-0.1in}
\begin{figure}[H]
    \centering
    \includegraphics[width=\columnwidth]{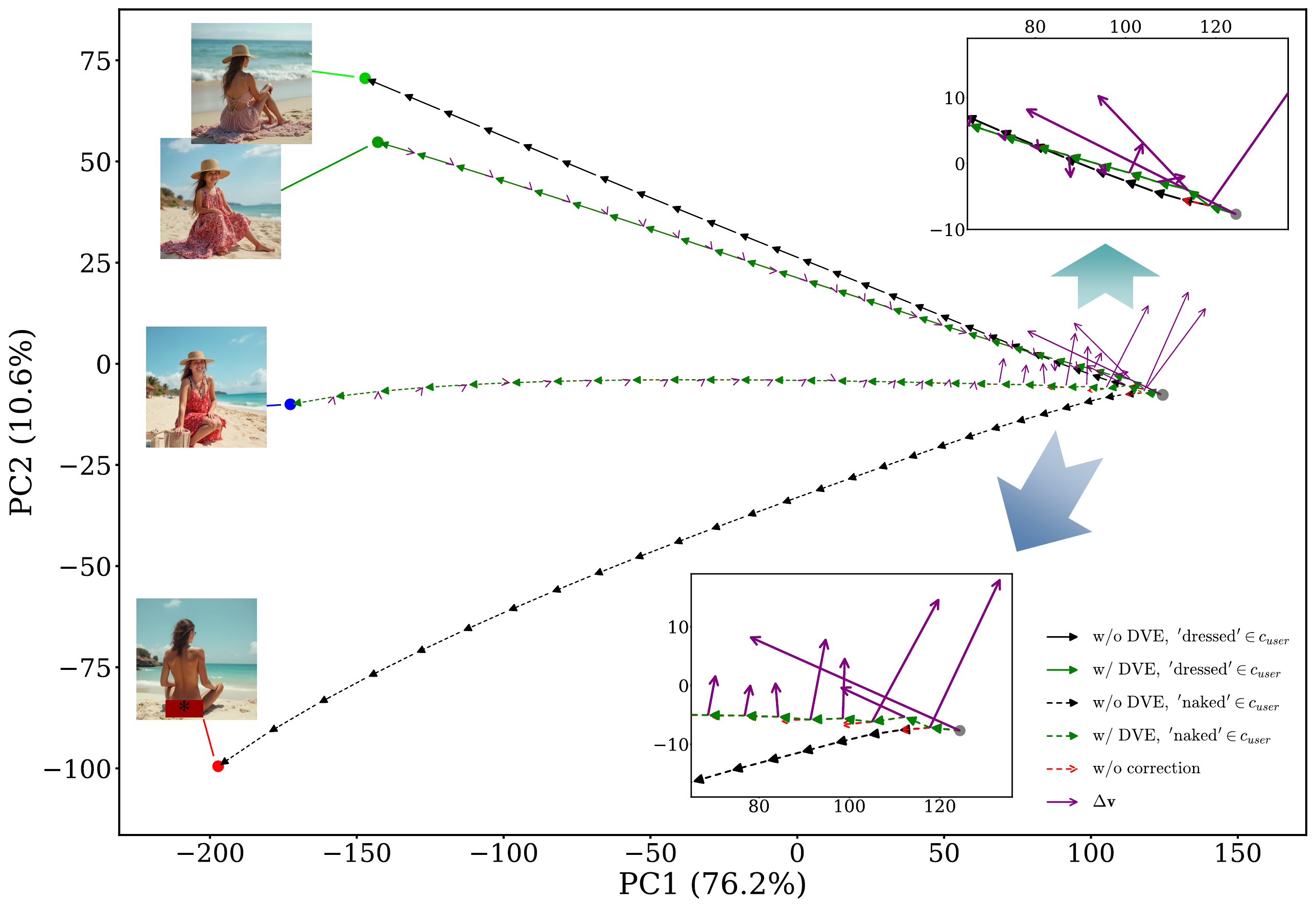}
    \caption{\textbf{PCA visualization of ODE trajectories.}}
    \label{fig:vis_tra}
\end{figure}
\end{minipage}

\vspace{-0.8in}
\end{wrapfigure}

We report UA measuring classification accuracy on erasure concepts, and IRA measuring accuracy on irrelevant concepts. \cref{tab:object} presents the results. DVE achieves the best UA and IRA, demonstrating effective erasure without degrading irrelevant concepts generation.

\paragraph{Artistic Style Erasure.}
We evaluate erasure of three famous artists' styles: Van Gogh, Leonardo da Vinci, and Pablo Picasso. 
We use a completely fixed template for all style erasure experiments: ``a painting by \textless artist\textgreater'' as the erasure prompt and ``a painting'' as the anchor prompt.
We use 50 prompts sourced from Concept-prune~\cite{conceptprune} and employ the style classifier from UnlearnDiffAtk~\cite{unlearndiffatk} to classify the generated images and calculate Top-1/3/5 accuracy. We also evaluate FID and CLIP on MSCOCO using 480 generated images for each artist's style and each method.

\cref{tab:style} shows that DVE achieves strong erasure performance across all three artists with competitive accuracy. The results demonstrate effective style removal while preserving painting quality.

\subsection{Qualitative Analysis}
\label{sec:qualitative_analysis}

To provide geometric intuition for DVE's erasure mechanism, we visualize the ODE trajectories in the latent space using Principal Component Analysis (PCA). We generate images under four configurations, with/without DVE and with prompts ``a dressed woman sitting on a beach'' or ``a naked woman sitting on a beach'', all starting from a same initial noise. We collect the latent states at each timestep along with the displacement endpoints, and project them onto the top-2 principal components.

\cref{fig:vis_tra} shows the projected trajectories and differential vectors. Without DVE, the ``naked'' trajectory diverges significantly downward from the ``dressed'' trajectory, terminating at a point corresponding to NSFW content. With DVE enabled, the ``naked'' trajectory is redirected upward by the differential vectors, converging toward the ``dressed'' trajectory and producing safe content. Meanwhile, the ``dressed'' trajectories with and without DVE remain nearly overlapping, demonstrating that DVE selectively corrects only concept-aligned velocities without affecting irrelevant generations. It also reveals that the differential vectors exhibit a consistent directional tendency across timesteps.

\subsection{Ablation Studies}

\paragraph{Impact of erasure strength and threshold.}
We conduct ablation studies to analyze the effect of key components. \cref{tab:abl} shows that without selective correction, CLIP score drops significantly, indicating over-erasure. Selective correction maintains CLIP scores while achieving effective erasure.

\begin{table*}
\centering
\begin{minipage}[t]{0.40\textwidth}
    \begin{table}[H]
    \caption{\textbf{Impact of erasure strength $\gamma$ and threshold $\tau$}. "psc-off" means not using projection-based selective correction. Best results are marked in \textbf{Bold}.}
    \label{tab:abl}
    \centering
    \small
    \begin{tabular}{cc|ccc}
        \toprule
        $\gamma$ & $\tau$ & Total$\downarrow$ & CLIP$\uparrow$ & FID$\downarrow$ \\
        \midrule
        \multicolumn{2}{c|}{Flux.1-dev} & 605 & 30.87 & 21.32 \\
        \midrule
        2 & psc-off & \textbf{22} & 24.83 & 41.54 \\
        2 & 0 & 155 & \textbf{30.18} & 21.91 \\
        3 & 0 & 52 & 28.60 & 23.66 \\
        \rowcolor{gray!25}
        3 & -15 & 146 & 30.04 & \textbf{21.70} \\
        \bottomrule
    \end{tabular}
    \end{table}
\end{minipage}
\hfill
\begin{minipage}[t]{0.58\textwidth}
    \begin{table}[H]
    \caption{\textbf{Computational cost reduction.} Comparison of erasure effectiveness and efficiency. The results shown are testing from I2P4703 and MSCOCO10k datasets. Latency denotes the average time to generate a single image. Best results are marked in \textbf{Bold}.}
    \label{tab:ccd}
    \centering
    \small
    \begin{tabular}{c|cccc}
        \toprule
        Method & Total$\downarrow$ & CLIP$\uparrow$ & FID$\downarrow$ & Latency(s)$\downarrow$ \\
        \midrule
        Flux.1-dev & 605 & 30.87 & 21.32 & 5.45 \\
        \midrule
        \rowcolor{gray!25}
        Ours & \textbf{146} & 30.04 & \textbf{21.70} & 16.37 \\
        + preprocessed & 226 & 29.45 & 25.27 & \textbf{5.49} \\
        + early-stage & 212 & \textbf{30.32} & 22.22 & 8.56 \\
        \bottomrule
    \end{tabular}
    \end{table}
\end{minipage}
\end{table*}

\begin{wrapfigure}{r}{0.55\textwidth}

\vspace{-0.4in}
\begin{minipage}[t]{0.55\textwidth}

\begin{table}[H]
    \caption{\textbf{Multi-concept erasure with 3 concepts per group}. The CLIP and FID shown are testing from MSCOCO480 dataset. Best results are marked in \textbf{Bold}.}
    \label{tab:multi}
    \centering
    \footnotesize
    \begin{tabular}{lcccc}
        \toprule
        Method & UA(\%)$\downarrow$ & IRA(\%)$\uparrow$ & CLIP$\uparrow$ & FID$\downarrow$ \\
        \midrule
        Flux.1-dev & \multicolumn{2}{c}{88.3} & 30.85 & 115.27 \\
        CA (Model-base) & \textbf{2.1} & 59.3 & 29.66 & 113.11 \\
        ESD & 23.9 & 67.2 & \textbf{30.46} & 115.15 \\
        EraseAnything & 23.1 & 55.6 & 29.32 & 116.51 \\
        \rowcolor{gray!25}
        \textbf{Ours} & 3.5 & \textbf{71.7} & 29.33 & \textbf{106.94} \\
        \bottomrule
    \end{tabular}
\end{table}

\vspace{-0.1in}

\begin{figure}[H]
    \centering
    \includegraphics[width=\columnwidth]{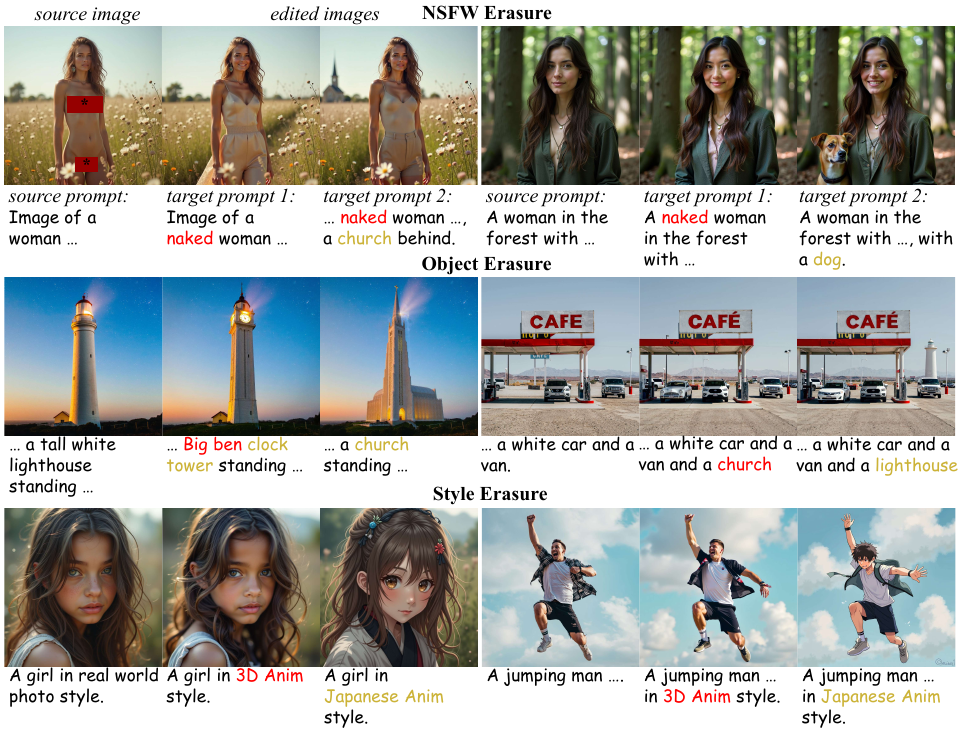}
    \caption{\textbf{Concept Erasure for Image Editing.}}
    \label{fig:exp_edit}
\end{figure}

\end{minipage}

\vspace{-0.4in}
\end{wrapfigure}

It demonstrates how erasure strength affects erasure intensity, and how a negative threshold helps preserve irrelevant concepts and image quality.

\paragraph{Computational Cost Reduction.} \label{par:computational_cost_reduction}
We evaluate the two proposed computational cost reduction methods. For the preprocessed differential vector, we use $M=2$, $\gamma=6$, and $\tau=-10$. For early-stage correction, we set $n^*=20$, applying corrections only during the first 8 steps. Results are shown in \cref{tab:ccd}. Early-stage correction substantially reduces computational overhead while largely preserving both erasure effectiveness and image quality. The preprocessed differential vector achieves even greater efficiency gains, 
though with some degradation in image quality.

\paragraph{Concept Erasure for Image Editing.}
We extend DVE to image editing via the FlowEdit~\cite{flowedit} framework. 
It requires two additional parameters: the lower and upper bounds of FlowEdit activate range $(n_{\min},n_{\max}]$.
Following its default settings, we set $n_{\max}=24$, with $n_{\min}=18$ for style edits and $n_{\min}=0$ for other edits.
DVE is not applied during the several time steps that affect coarse-grained modifications in the range $n_{\max}+1$ to $T$. As shown in \cref{fig:exp_edit}, DVE effectively rejects the generation of erasure concepts while preserving irrelevant content, and is capable of erasing concepts already inherent in the source image.

\paragraph{Multi-Concept Erasure.}
We also evaluate multi-concept erasure using 12 groups of 3 object concepts each. Results are shown in \cref{tab:multi}. DVE achieves competitive UA while obtaining the best IRA. 
More details are shown in \cref{tab:multi_detail}.

\section{Conclusion}
\label{sec:conclusion}

We presented DVE, a training-free concept erasure method tailored for flow matching models. By introducing the differential vector field and projection-based selective correction, DVE achieves precise concept erasure while preserving the overall generation ability of the text-to-image model. Experimental results demonstrate that our method achieves state-of-the-art performance across various tasks including both T2I generation and editing. Requiring zero training, DVE is instantly applicable to flow matching models, providing a practical solution for responsible AI deployment.

\begin{ack}
This work is supported in part by the National Natural Science Foundation of China under grants 62301189, 62576122, and 62571298, Guangdong Basic and Applied Basic Research Foundation under grant 2026A1515011139. The authors declare no competing interests.
\end{ack}

\newpage

{
\small

\bibliography{main}
\bibliographystyle{unsrt}
}


\newpage
\appendix

\section*{Appendix Overview}

This appendix provides supplementary materials to support the main paper as follows:

\begin{itemize}
    \item \textbf{Theoretical Analysis} (\cref{sec:theory_appendix}): provides geometric analysis of the projection-based selective correction mechanism and semantic subspace confinement, establishing theoretical justification for DVE's selective erasure property.

    \item \textbf{Algorithm Details} (\cref{sec:alg_details}): presents the complete algorithms for multi-concept erasure and image editing with DVE.

    \item \textbf{Implementation Details} (\cref{sec:impl_details}): details the prompts and unified anchor concept selection rules used in different domains, the hyper-parameters for all experiments, and the evaluation metrics.

    \item \textbf{Additional Results} (\cref{sec:add_results}): presents additional experimental results, including comparison with more diffusion-based inference-time baselines on NSFW erasure, per-concept results for object erasure, LLM-generated anchor concept experiments, extension to SD3.5 Medium.

    \item \textbf{Supplementary Explanation} (\cref{sec:e_supplementary_explanation}): explains the fundamental differences between DVE and other diffusion-based inference-time methods, including GLoCE, ActErase, SAFREE, Semantic Surgery and SLD, and explains why Negative Prompt is not suitable for concept erasure, and discusses the Basis for using MSCOCO480.

    \item \textbf{Limitations} (\cref{sec:limitations}): discusses the limitations of DVE regarding its applicability to the SDE paradigm and the trade-offs introduced by the proposed computational cost reduction strategies.
    
    \item \textbf{Broader Impacts} (\cref{sec:broader_impacts}): discusses the potential positive and negative societal impacts of DVE for the responsible deployment of generative AI.
\end{itemize}
\section{Theoretical Analysis}
\label{sec:theory_appendix}

\subsection{Geometric Analysis of Selective Erasure}
\label{sec:theory}

We provide theoretical justification for the projection-based correction mechanism. The key insight is that the projection score $\st$ geometrically encodes the relationship between the user's velocity and the concept direction, as illustrated in \cref{fig:schematic_totalmethod}.

Let $\theta$ denote the angle between $\vt_{\text{user}}$ and $\deltav$. 
Then the projection score can be calculated by $\st = \|\vt_{\text{user}}\| \cos\theta$. 
Since $\deltav = \vt_{\text{anc}} - \vt_{\text{era}}$ points from the erasure concept $\cera$ toward the anchor concept $\canc$, the sign of $\cos\theta$ indicates whether the user's velocity aligns with or opposes the safe direction. The threshold parameter $\tau$ should be set to a value less than or equal to 0; the rationale for this choice will become clear from the case analysis below. We analyze three representative cases to illustrate how projection-based selective correction works:

\paragraph{Case 1: Prompt contains the erasure concept.}
When the user prompt contains the erasure concept (e.g., ``a naked woman''), the user velocity $\vt_{\text{user}}$ points toward $\cera$, opposite to $\deltav$, yielding $\theta > 90^\circ$. Consequently, a correction with magnitude $\gamma(\tau - \st)$ is applied to redirect generation toward the safe direction. Setting $\gamma > 1$ enables stronger erasure: when the generated content contains the erasure concept, the correction pushes the trajectory further toward the anchor concept distribution, effectively preventing the generation of undesirable content while minimally affecting irrelevant concepts.

\paragraph{Case 2: Prompt contains the anchor concept.}
When the user prompt contains the anchor concept (e.g., ``a dressed woman''), $\vt_{\text{user}}$ aligns with $\deltav$, which satisfies $\theta < 90^\circ$. In this case $\st > \tau$, so no correction is applied.

\paragraph{Case 3: Prompt contains irrelevant concepts.}
When the user prompt is irrelevant to both concepts (e.g., ``an apple''), ideally $\vt_{\text{user}}$ and $\deltav$ would be orthogonal, yielding no correction. However, the velocity space of flow models is not a perfectly idealized semantic space, so in practice $\st$ will have a small absolute value. We therefore set $\tau < 0$ to leave this case without correction as much as possible, minimizing unintended effects on irrelevant content. Additionally, setting $\gamma > 1$ compensates for the reduced erasure strength in Case 1 caused by the negative threshold.

This geometric analysis establishes that projection-based selective correction achieves \emph{selective erasure}: correction is applied only when the velocity has negative alignment with $\deltav$, while orthogonal or positively aligned velocities remain untouched.

\subsection{Semantic Subspace Confinement}
\label{sec:subspace}

Beyond selectivity, we analyze whether corrections introduce unintended side effects. A natural concern is that modifying the velocity field might push the latent trajectory off the data manifold, degrading image quality. We show that DVE corrections are minimal in a precise sense.

Recent work on diffusion model editing~\cite{locoedit} reveals that the Jacobian of the velocity field $J_t = \nabla_{\zt} \vt(\zt_t, t, c)$ exhibits low numerical rank: $\text{rank}(J_t) = r \ll d$ where $d$ is the latent dimension. This defines two orthogonal subspaces: the semantic subspace $\mathcal{C}_t = \text{range}(J_t)$ containing directions that affect generation semantics, and the nullspace $\mathcal{N}_t = \mathcal{C}_t^\perp$ containing directions with no semantic effect.

We assume that the differential vector field lies in the semantic subspace, i.e. $\deltav(\zt_t, t) \in \mathcal{C}_t$, since the difference between ``nudity'' and ``dressed'' is a semantic distinction.

Let $\zt_t$ denote the original trajectory following $\frac{d\zt_t}{dt} = -\vt(\zt_t, t, \cuser)$, and let $\tilde{\zt}_t$ denote the corrected trajectory:
\begin{equation}
    \frac{d\tilde{\zt}_t}{dt} = -(\vt(\tilde{\zt}_t, t, \cuser) + \delta\vt_t),
    \label{eq:corrected_ode}
\end{equation}
where the correction term is $\delta\vt_t = \gamma(\tau - \st) \cdot \mathbf{1}[\st < \tau] \cdot \frac{\deltav}{\|\deltav\|}$. Define the trajectory deviation $e_t = \tilde{\zt}_t - \zt_t$. Taking the difference of the two ODEs and applying Taylor expansion:
\begin{align}
    \frac{de_t}{dt} &= \frac{d\tilde{\zt}_t}{dt} - \frac{d\zt_t}{dt} \notag \\
    &= -(\vt(\tilde{\zt}_t, t, \cuser) + \delta\vt_t - \vt(\zt_t, t, \cuser) \notag) \\
    &= -(\vt(\zt_t, t, \cuser) + J_t e_t + O(\|e_t\|^2) + \delta\vt_t - \vt(\zt_t, t, \cuser) \notag) \\
    &= -(J_t e_t + \delta\vt_t + O(\|e_t\|^2)).
    \label{eq:deviation_ode}
\end{align}
Projecting onto the nullspace $\mathcal{N}_t$, we have $P_{\mathcal{N}_t}[J_t e_t] = 0$ since $J_t e_t \in \text{range}(J_t) = \mathcal{C}_t$, and $P_{\mathcal{N}_t}[\delta\vt_t] = 0$ since $\delta\vt_t \propto \deltav \in \mathcal{C}_t$ by assumption. Therefore, the nullspace component of the deviation satisfies:
\begin{equation}
    \frac{d}{dt}P_{\mathcal{N}_t}[e_t] = O(\|e_t\|^2).
    \label{eq:nullspace_result}
\end{equation}
This shows that the nullspace component has no first-order driving term. DVE corrections primarily affect the semantic subspace $\mathcal{C}_t$, with only higher-order effects on the nullspace. Our method thus achieves minimal sufficient correction: modifying only what is necessary for erasure while preserving image quality.

This property is crucial because although the nullspace $\mathcal{N}_t$ corresponds to non-semantic dimensions, excessive nullspace deviation could push the latent toward low-density regions, cause the decoder to produce low-quality images, or introduce high-frequency artifacts and texture loss.

\section{Algorithm Details}
\label{sec:alg_details}

\subsection{Multi-Concept Erasure}
\label{sec:alg_multi}

\cref{alg:DVE} in the main paper presents the single-concept version of DVE for clarity. Here we provide the complete multi-concept extension that supports simultaneous erasure of multiple concepts.

For multi-concept erasure, each concept is processed independently: we compute the projection score $\st^{(i)}$ for each concept $i$ separately, and apply correction only for those concepts where $\st^{(i)} < \tau$. This independent projection approach ensures that erasure of one concept does not interfere with the preservation of irrelevant concepts. See~\cref{alg:DVE_multi} for details.

\begin{algorithm}[htbp!]
    \caption{Multi-Concept Erasure DVE}
    \label{alg:DVE_multi}
    \begin{algorithmic}
        \STATE \textbf{Input} user prompt $\cuser$, erasure prompts, anchor prompts and erasure strengths $\{(\cera^{(i)}, \canc^{(i)}, \gamma_i)\}_{i=1}^{k}$, threshold $\tau$, total steps $T$
        \STATE \textbf{Output} generated image $\xt_0$
        \STATE Initialize $\zt_1 \sim \mathcal{N}(0, \mathbf{I})$
        \FOR{$n = T$ to $1$}
            \STATE $t_n \leftarrow n / T$
            \STATE $\vt_{\text{user}} \leftarrow \vt(\zt_{t_n}, t_n, \cuser)$
            \FOR{$i = 1$ to $k$}
                \STATE $\deltav^{(i)} \leftarrow \vt(\zt_{t_n}, t_n, \canc^{(i)}) - \vt(\zt_{t_n}, t_n, \cera^{(i)})$
                \STATE $\st^{(i)} \leftarrow \vt_{\text{user}} \cdot \frac{\deltav^{(i)}}{\|\deltav^{(i)}\|}$
            \ENDFOR
            \STATE $\vt_{\text{corr}} \leftarrow \vt_{\text{user}} + \sum_{i: \st^{(i)} < \tau} \gamma_i (\tau - \st^{(i)}) \cdot \frac{\deltav^{(i)}}{\|\deltav^{(i)}\|}$
            \STATE $\zt_{t_{n-1}} \leftarrow \zt_{t_n} + (t_n - t_{n-1}) \cdot \vt_{\text{corr}}$
        \ENDFOR
        \STATE \textbf{return} $\text{Decode}(\zt_0)$
    \end{algorithmic}
\end{algorithm}

\begin{algorithm}[htbp!]
    \caption{Image Editing with DVE}
    \label{alg:DVE_edit}
    \begin{algorithmic}
        \STATE \textbf{Input} source image $\xt^{\text{src}}$, source prompt $c^{\text{src}}$, target prompt $c^{\text{tar}}$, erasure concept $c^{\text{era}}$, anchor concept $c^{\text{anc}}$, erasure strength $\gamma$, threshold $\tau$, FlowEdit active range $(n_{\min}, n_{\max}]$, total steps $T$
        \STATE \textbf{Output} edited image $\xt_0$
        \STATE $\zt^{\text{edit}}_0 \leftarrow \text{Encode}(\xt^{\text{src}})$
        \FOR{$n = n_{\max}$ to $n_{\min} + 1$}
            \STATE $t_n \leftarrow n/T$
            \STATE $\epsilon \sim \mathcal{N}(0, \mathbf{I})$
            \STATE $\zt^{\text{src}}_{t_n} \leftarrow (1-t_n) \cdot \xt^{\text{src}} + t_n \cdot \epsilon$
            \STATE $\zt^{\text{tar}}_{t_n} \leftarrow \zt^{\text{edit}}_{t_n} + \zt^{\text{src}}_{t_n} - \xt^{\text{src}}$
            \STATE $\vt^{\text{tar}} \leftarrow \vt(\zt^{\text{tar}}_{t_n}, t_n, c^{\text{tar}})$
            \STATE $\deltav \leftarrow \vt(\zt^{\text{tar}}_{t_n}, t_n, c^{\text{anc}}) - \vt(\zt^{\text{tar}}_{t_n}, t_n, c^{\text{era}})$
            \STATE $\st \leftarrow \vt^{\text{tar}} \cdot \frac{\deltav}{\|\deltav\|}$
            \STATE $\vt^{\text{corr}} \leftarrow \vt^{\text{tar}} + \mathbf{1}[\st < \tau] \cdot \gamma (\tau - \st) \cdot \frac{\deltav}{\|\deltav\|}$
            \STATE $\zt^{\text{edit}}_{t_{n-1}} \leftarrow \zt^{\text{edit}}_{t_n} + (t_n - t_{n-1}) \cdot (\vt^{\text{corr}} - \vt(\zt^{\text{src}}_{t_n}, t_n, c^{\text{src}}))$
        \ENDFOR
        \IF{$n_{\min} > 0$}
            \STATE $\epsilon \sim \mathcal{N}(0, \mathbf{I})$
            \STATE $\zt^{\text{tar}}_{t_{n_{\min}}} \leftarrow \zt^{\text{edit}}_{t_{n_{\min}}} + t_{n_{\min}} \cdot \epsilon + (1 - t_{n_{\min}}) \cdot \zt^{\text{src}}_0 - \zt^{\text{src}}_0$
            \FOR{$n = n_{\min}$ to $1$}
                \STATE $t_n \leftarrow n/T$
                \STATE $\vt^{\text{tar}} \leftarrow \vt(\zt^{\text{tar}}_{t_n}, t_n, c^{\text{tar}})$
                \STATE $\deltav \leftarrow \vt(\zt^{\text{tar}}_{t_n}, t_n, c^{\text{anc}}) - \vt(\zt^{\text{tar}}_{t_n}, t_n, c^{\text{era}})$
                \STATE $\st \leftarrow \vt^{\text{tar}} \cdot \frac{\deltav}{\|\deltav\|}$
                \STATE $\vt^{\text{corr}} \leftarrow \vt^{\text{tar}} + \mathbf{1}[\st < \tau] \cdot \gamma (\tau - \st) \cdot \frac{\deltav}{\|\deltav\|}$
                \STATE $\zt^{\text{tar}}_{t_{n-1}} \leftarrow \zt^{\text{tar}}_{t_n} + (t_n - t_{n-1}) \cdot \vt^{\text{corr}}$
            \ENDFOR
        \ENDIF
        \STATE \textbf{return} $\text{Decode}(\zt^{\text{edit}}_0)$ if $n_{\min} = 0$ else $\text{Decode}(\zt^{\text{tar}}_0)$
    \end{algorithmic}
\end{algorithm}

\subsection{Image Editing with DVE}
\label{sec:alg_edit}

FlowEdit~\cite{flowedit} constructs direct ODE paths for image editing. Given source image $\xt^{\text{src}}=\zt^{\text{src}}_{0}=\zt^{\text{edit}}_{0}$, source prompt $c^{\text{src}}$ and target prompt $c^{\text{tar}}$, it defines an editing trajectory $\zt^{\text{edit}}_t$ with velocity:
\begin{equation}
    \frac{d\zt^{\text{edit}}_t}{dt} = \vt(\zt^{\text{tar}}_t, t, c^{\text{tar}}) - \vt(\zt^{\text{src}}_t, t, c^{\text{src}}),
    \label{eq:flowedit}
\end{equation}
where $\zt^{\text{tar}}_t = \zt^{\text{edit}}_t + \zt^{\text{src}}_t - \zt^{\text{src}}_0$. This enables structure-preserving edits by transporting along velocity differences.

DVE naturally integrates with flow-based image editing frameworks such as FlowEdit. DVE correction is applied to the target velocity $\vt^{\text{tar}}$, consistent with how we correct $\vt_{\text{user}}$ in T2I generation. This ensures that undesirable concepts are suppressed regardless of whether they appear in the target prompt or are inherent in the source image. See~\cref{alg:DVE_edit} for details.

\section{Implementation Details}
\label{sec:impl_details}

\subsection{Details on Prompts}
\label{sec:c_details_on_prompts}

DVE requires specifying an erasure concept and an anchor concept to compute the differential vector field. \cref{tab:prompts} shows the erasure concepts we use in different domains, the selected anchor concepts and the formulation of erasure and anchor prompts for different domains. 

Regarding the rules for selecting the anchor concept: we prefer to use the generalized form of the erasure concept as the anchor concept. If it is difficult to find such a generalized concept, we instead select a concept that is similar but clearly distinct. Similar rules for selecting anchor concepts have been adopted in previous work~\cite{ca}. Experiments have shown that this rule for selecting anchor concepts can retain the generation effect of irrelevant content while erasing the erasure concept.

\begin{table}[h]
\centering
\caption{Details of the erasure concepts we use in different domains, the selected anchor concepts and the formulation of erasure and anchor prompts for different domains. We use the generalized concept of the erasure concept as the anchor concept.}
\label{tab:prompts}
\small
\setlength{\tabcolsep}{5pt}
    \begin{tabular}{c|cc|cc}
    \toprule
    Domain & Erasure Concept & Anchor Concept & Erasure Prompt & Anchor Prompt \\
    \midrule
    NSFW & nudity & dressed & \textless erasure concept\textgreater & \textless anchor concept\textgreater \\
    \midrule
    \multirow{12}{*}{Object} & church & building & \multirow{12}{*}{\textless erasure concept\textgreater} & \multirow{12}{*}{\textless anchor concept\textgreater} \\
     & golf ball & ball & & \\
     & English springer & dog & & \\
     & cassette player & electronic device & & \\
     & chain saw & tool & & \\
     & French horn & instrument & & \\
     & garbage truck & truck & & \\
     & gas pump & pump & & \\
     & parachute & cloth & & \\
     & banana & fruit & & \\
     & Siamese cat & cat & & \\
     & goose & bird & & \\
    \midrule
    \multirow{3}{*}{Art Style} & Vincent van Gogh & - & \multirow{3}{*}{a painting by \textless erasure concept\textgreater} & \multirow{3}{*}{a painting} \\
     & Leonardo Da Vinci & - & & \\
     & Pablo Picasso & - & & \\
    \bottomrule
    \end{tabular}
\end{table}

To further demonstrate the effectiveness of the rules and the robustness of the anchor prompt, we also conducted experiments on LLM-generated anchor concepts, see~\cref{sec:d_llm_generated_anchor_concept}.

\subsection{Details on Hyper-parameters}
\label{sec:c_details_on_hyper_parameters}

\cref{tab:hyperparams} lists the hyper-parameters used in our experiments. The same parameters were used for all T2I task experiments. For the $n_{\max}$ and $n_{\min}$ of the edit task, we used the parameters recommended by FlowEdit~\cite{flowedit}.

\begin{table}[h]
\centering
\caption{Details on hyper-parameters, including erasure strength $\gamma$ and threshold $\tau$ for projection-based selective correction for different tasks.}
\label{tab:hyperparams}
    \begin{tabular}{cc|cc|cc}
    \toprule
    Task & Domain & $\gamma$ & $\tau$ & $n_{\max}$ & $n_{\min}$ \\
    \midrule
    \multirow{3}{*}{T2I} & NSFW & 3.0 & -15 & - & - \\
     & Object & 3.0 & -15 & - & - \\
     & Style & 3.0 & -15 & - & - \\
    \midrule
    \multirow{3}{*}{Edit} & NSFW & 3.0 & 0 & 24 & 0 \\
     & Object & 3.0 & 0 & 24 & 0 \\
     & Style & 3.0 & 0 & 24 & 18 \\
    \bottomrule
    \end{tabular}
\end{table}

\subsection{Details on Evaluation Metrics}
\label{appendix:metrics}

We introduce the evaluation metrics used in our experiments. We calculate different metrics for each task to comprehensively evaluate both erasure effectiveness and generation quality.

\paragraph{NudeNet.}
For NSFW content erasure, we use NudeNet~\cite{nudenet} detector with threshold 0.6 to identify exposed body parts in generated images. We count the total number of detected exposed body part bounding boxes across all generated images. A lower total count indicates more effective nudity erasure. We also calculate ASR for the images generated on the adversarial attack datasets, using the ratio of images detected with exposed body parts and the total number of images generated.

\paragraph{UA and IRA.}
For object and style erasure tasks, we report Unlearning Accuracy (UA) and In-domain Retain Accuracy (IRA). For all concepts selected under a domain, UA measures the classification accuracy on prompts containing the erasure concept, where a lower UA indicates more effective erasure. IRA measures the classification accuracy on prompts containing irrelevant concepts from the same domain, where a higher IRA indicates better preservation of irrelevant content. For object erasure, we use ResNet-50~\cite{resnet} pretrained on ImageNet~\cite{imagenet} as the classifier. For style erasure, we use the classifier from UnlearnDiffAtk~\cite{unlearndiffatk} and report Top-1/3/5 accuracy.

\begin{figure}[t!]
    \centering
    \includegraphics[width=\columnwidth]{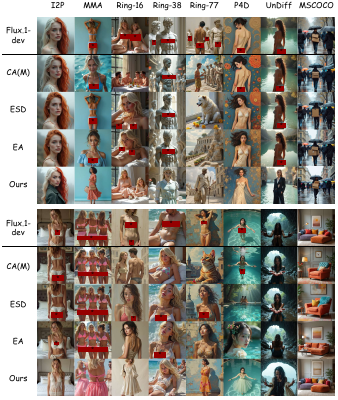}
    \caption{Additional visualization results for nudity erasure on each adversarial attack dataset.}
    \label{fig:add_advatk}
\end{figure}

\paragraph{FID.}
Fr\'{e}chet Inception Distance measures the quality and diversity of generated images by comparing their feature statistics to real images. The FID is calculated as:
\begin{equation}
    \text{FID} = \|\mu_r - \mu_g\|^2 + \operatorname{Tr}\left(\Sigma_r + \Sigma_g - 2\sqrt{\Sigma_r \Sigma_g}\right)
\end{equation}
where $\mu_r$, $\mu_g$ and $\Sigma_r$, $\Sigma_g$ are the mean and covariance of the real and generated features extracted by Inception-v3 network, respectively. A lower FID indicates better image quality. We use the MSCOCO30k~\cite{mscoco} validation set as reference images.

\paragraph{CLIP Score.}
CLIP Score measures the semantic alignment between generated images and their corresponding text prompts. It uses the CLIP model~\cite{clip} to project images and text into a shared embedding space, and computes the cosine similarity between embeddings. A higher CLIP Score indicates stronger text-image correspondence, meaning the generated image better reflects the content described in the prompt.
\section{Additional Results}
\label{sec:add_results}

\subsection{Additional Main Results}
\label{sec:add_main}

\paragraph{NSFW Erasure.} \label{par:d_nsfw_erasure}
We present additional visualization results for nudity erasure on each dataset in \cref{fig:add_advatk}.

\begin{figure}[h!]
    \centering
    \includegraphics[width=\columnwidth]{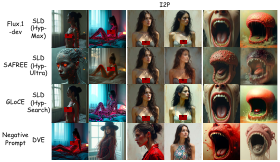}
    \caption{Additional visualization results for more inference-time concept erasure methods originally designed for diffusion models.}
    \label{fig:add_more_inference_time_baseline}
\end{figure}

Our method is a training-free concept erasure approach specifically designed for flow matching models, and is fundamentally different from inference-time methods developed for diffusion models. To provide a more comprehensive comparison, we migrate a broader set of inference-time diffusion model methods to flow matching models. While some methods are heavily coupled with diffusion model architectures and are difficult or even infeasible to transfer, others with weaker architectural dependencies can be successfully migrated. We additionally include negative prompt, a technique not commonly used for concept erasure in image generation, as a reference baseline. As shown in \cref{tab:inference_time_nsfw}, our method significantly outperforms these diffusion-model inference-time methods, demonstrating the unique advantage of DVE for flow matching models. For a visual comparison of DVE and various inference-time erasure methods for diffusion models, see~\cref{fig:add_more_inference_time_baseline}.

\begin{table}[h!]
    \caption{Comparison with inference-time methods on NSFW erasure. We report total exposed body parts count on I2P4703 and image quality metrics on MSCOCO480. Best results are marked in \textbf{Bold}.}
    \label{tab:inference_time_nsfw}
    \centering
    \begin{tabular}{lccc}
        \toprule
        Method & Total$\downarrow$ & CLIP$\uparrow$ & FID$\downarrow$ \\
        \midrule
        w/o unlearn & 605 & 30.85 & 115.27 \\
        SAFREE & 440 & 28.61 & 113.73 \\
        GLoCE & 512 & 30.42 & 115.95 \\
        Negative Prompt & 163 & 29.85 & 116.12 \\
        SLD (Hyp-Max) & 362 & 31.61 & 113.82 \\
        SLD (Hyp-Ultra) & 512 & 30.19 & 112.15 \\
        SLD (Hyp-Search) & 218 & \textbf{31.83} & 117.44 \\
        DVE & \textbf{146} & 29.95 & \textbf{108.99} \\
        \bottomrule
    \end{tabular}
\end{table}

\begin{table}[h!]
    \caption{Comparison with negative prompt on object erasure. Best results among unlearning methods are marked in \textbf{Bold}.}
    \label{tab:neg_prompt_object}
    \centering
    \begin{tabular}{lcccc}
        \toprule
        Method & UA(\%)$\downarrow$ & IRA(\%)$\uparrow$ & CLIP$\uparrow$ & FID$\downarrow$ \\
        \midrule
        w/o unlearn & \multicolumn{2}{c}{88.3} & 30.85 & 115.27 \\
        Negative Prompt & 82.9 & 71.7 & 30.05 & 120.66 \\
        DVE & \textbf{3.3} & \textbf{86.3} & \textbf{30.43} & \textbf{112.63} \\
        \bottomrule
    \end{tabular}
\end{table}

For all methods, we use their default parameters. For SLD, we first tried its official configurations; almost all produce negligible erasure except Hyp-Max, the strongest, which still shows limited effect. Since Hyp-Max already exceeds the range recommended in the original SLD paper, we explored beyond that range with $s_S = 20000$, $\lambda = 2.0$, $\delta = 0$, $s_m = 1.0$, $\beta_m = 1.2$, denoted as Hyp-Ultra. However, an excessively large threshold causes almost all elements to be corrected, including those moving safely. We then searched and identified $s_S = 20000$, $\lambda = 0.005$, $\delta = 0$, $s_m = 1.0$, $\beta_m = 0.8$, denoted as Hyp-Search. Although its CLIP looks relatively high, when actually observing the images it generates, it is found that the image quality has seriously declined, with blurriness and artifacts. For a theoretical analysis of the differences between SLD and our method, see~\cref{par:e_sld}.

Regarding negative prompt, although its erasure of nudity on the I2P dataset is only slightly inferior to our method, it cannot generalize to diverse concept erasure tasks. First, it lacks any selective mechanism, the negative prompt is applied unconditionally at every step, which degrades the generation quality of irrelevant content. Second, when the positive prompt is relatively short and explicitly contains the same concept as the negative prompt, the effect of the negative prompt is largely cancelled out by the positive prompt, making erasure ineffective. As shown in the object erasure experiment in \cref{tab:neg_prompt_object}, where prompts are short and explicitly contain the erasure concept, IRA barely decreases, while UA for several categories drops sharply due to blurriness and color distortion in generated images. For a theoretical analysis of the Negative Prompt's limitations, see~\cref{par:e_negative_prompt}.

\paragraph{Object Erasure.}

\cref{tab:object_detail} shows per-concept results, where DVE achieves 0\% UA on 7 out of 12 concepts.

\begin{table}[h!]
    \caption{Per-concept UA and IRA for object erasure. Best results are marked in \textbf{Bold}. DVE achieves complete erasure on seven of these concepts.}
    \label{tab:object_detail}
    \centering
    \setlength{\tabcolsep}{3pt}
    \begin{tabular}{lccccccccc}
        \toprule
        \multirow{2}{*}{Concept} & ACC & \multicolumn{4}{c}{UA $\downarrow$} & \multicolumn{4}{c}{IRA $\uparrow$} \\
        \cmidrule(lr){2-2} \cmidrule(lr){3-6} \cmidrule(lr){7-10}
        & Flux.1-dev & CA(M) & ESD & EA & Ours & CA(M) & ESD & EA & Ours \\
        \midrule
        church & .70 & .25 & .15 & .80 & \textbf{.00} & .80 & .78 & \textbf{.88} & .84 \\
        golf ball & 1.00 & .15 & .30 & .15 & \textbf{.05} & .85 & .77 & .83 & \textbf{.88} \\
        English springer & .65 & .00 & .05 & .05 & \textbf{.00} & .87 & \textbf{.90} & .83 & .89 \\
        cassette player & .30 & .00 & .10 & .00 & \textbf{.00} & .87 & .93 & \textbf{.94} & .90 \\
        chain saw & 1.00 & .00 & .00 & \textbf{.00} & .05 & \textbf{.85} & .84 & .82 & .82 \\
        French horn & 1.00 & \textbf{.00} & .70 & .85 & .05 & .86 & .77 & .80 & \textbf{.86} \\
        garbage truck & 1.00 & .00 & \textbf{.00} & .30 & .15 & .79 & .84 & .79 & \textbf{.85} \\
        gas pump & 1.00 & \textbf{.00} & .10 & .05 & .10 & .86 & .84 & \textbf{.86} & .85 \\
        parachute & .95 & .00 & .00 & .00 & \textbf{.00} & .70 & .81 & .80 & \textbf{.89} \\
        banana & 1.00 & .10 & .95 & .75 & \textbf{.00} & .84 & .81 & .77 & \textbf{.85} \\
        Siamese cat & 1.00 & .00 & .15 & .00 & \textbf{.00} & .79 & .86 & .83 & \textbf{.92} \\
        goose & 1.00 & .00 & .65 & .50 & \textbf{.00} & .72 & \textbf{.85} & .80 & .80 \\
        \bottomrule
    \end{tabular}
\end{table}

\begin{table}[h!]
    \caption{Per-multi-concepts UA and IRA for 3-concept object erasure. Best results are marked in \textbf{Bold}.}
    \label{tab:multi_detail}
    \centering
    \setlength{\tabcolsep}{3pt}
    \footnotesize
    \begin{tabular}{lccccccccc}
        \toprule
        \multirow{2}{*}{Concepts} & ACC & \multicolumn{4}{c}{UA $\downarrow$} & \multicolumn{4}{c}{IRA $\uparrow$} \\
        \cmidrule(lr){2-2} \cmidrule(lr){3-6} \cmidrule(lr){7-10}
        & Flux.1-dev & CA(M) & ESD & EA & Ours & CA(M) & ESD & EA & Ours \\
        \midrule
        church + golf ball + English springer & .78 & .05 & .18 & .10 & \textbf{.03} & .67 & .73 & .64 & \textbf{.80} \\
        golf ball + English springer + cassette player & .65 & .03 & .30 & .12 & \textbf{.00} & .74 & .78 & .68 & \textbf{.86} \\
        English springer + cassette player + chain saw & .65 & \textbf{.00} & .02 & .10 & .05 & .84 & \textbf{.87} & .76 & .73 \\
        cassette player + chain saw + French horn & .77 & \textbf{.00} & .25 & .37 & .03 & .77 & \textbf{.84} & .71 & .61 \\
        chain saw + French horn + garbage truck & 1.00 & \textbf{.00} & .25 & .30 & .07 & \textbf{.72} & .71 & .50 & .64 \\
        French horn + garbage truck + gas pump & 1.00 & \textbf{.00} & .30 & .38 & .02 & \textbf{.68} & .66 & .57 & .67 \\
        garbage truck + gas pump + parachute & .98 & .00 & \textbf{.00} & .10 & .07 & .56 & .67 & .62 & \textbf{.72} \\
        gas pump + parachute + banana & .98 & .07 & .12 & .13 & \textbf{.07} & .51 & .53 & .41 & \textbf{.72} \\
        parachute + banana + Siamese cat & .98 & .02 & .20 & .12 & \textbf{.00} & .24 & .47 & .22 & \textbf{.78} \\
        banana + Siamese cat + goose & 1.00 & .02 & .68 & .35 & \textbf{.00} & .40 & .59 & .41 & \textbf{.71} \\
        Siamese cat + goose + church & .90 & \textbf{.02} & .35 & .32 & .05 & .47 & \textbf{.68} & .58 & .64 \\
        goose + church + golf ball & .90 & .05 & .22 & .38 & \textbf{.03} & .51 & .53 & .56 & \textbf{.72} \\
        \bottomrule
    \end{tabular}
\end{table}

\begin{figure}[h!]
    \centering
    \includegraphics[width=\columnwidth]{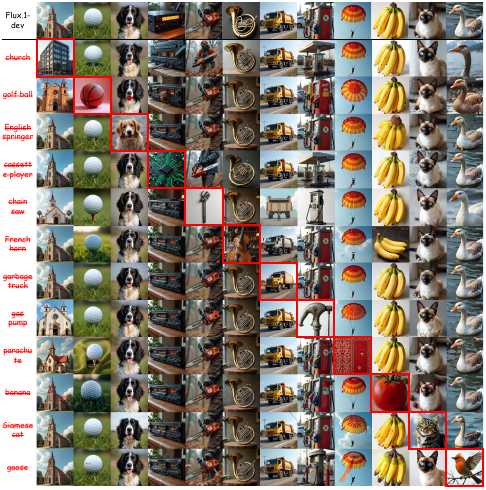}
    \caption{Additional visualization results for object erasure across erasure concepts and prompts used for generation. The image enclosed in the red box is the image containing the concept to be erased.}
    \label{fig:add_object}
\end{figure}

We also present additional visualization results for object erasure across erasure concepts and prompts used for generation using our method in \cref{fig:add_object}.

\begin{figure}[h!]
    \centering
    \includegraphics[width=0.75\columnwidth]{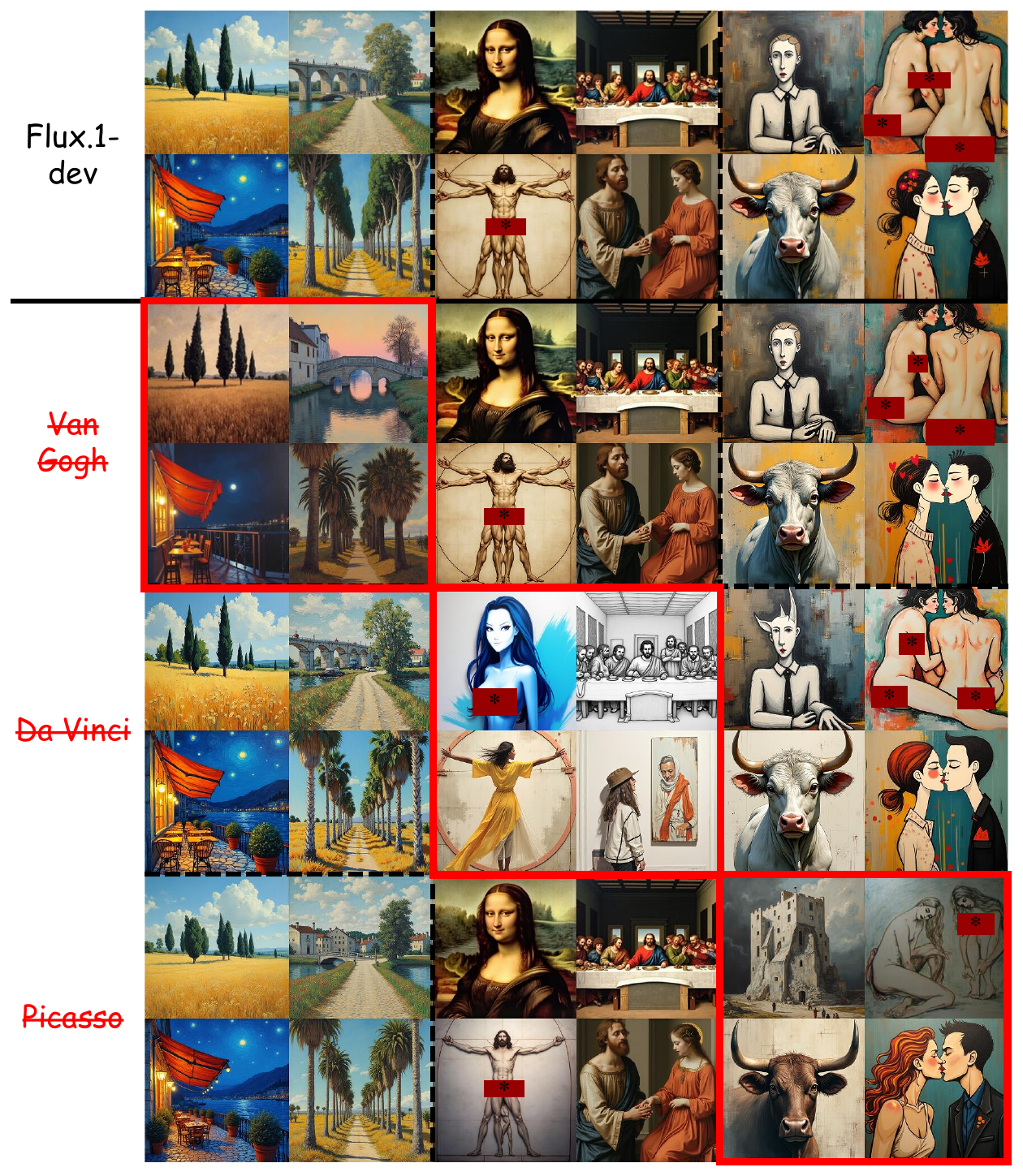}
    \caption{Additional visualization results for artist style erasure. The image enclosed in the red box is the image containing the concept to be erased.}
    \label{fig:add_style}
\end{figure}

\paragraph{Style Erasure.}
We present additional visualization results for artist style erasure using our method in \cref{fig:add_style}.

\subsection{Additional Ablation Study Results}
\label{sec:add_ablation}

\begin{figure}[h!]
    \centering
    \includegraphics[width=0.75\columnwidth]{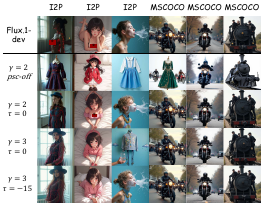}
    \caption{Additional visualization results about the ablation study of impact of erasure strength and threshold.}
    \label{fig:add_abl}
\end{figure}

\paragraph{Impact of erasure strength and threshold.}
We present additional visualization results about the ablation study of impact of erasure strength and threshold in \cref{fig:add_abl}.

\paragraph{Multi-Concept Erasure.}
\cref{tab:multi_detail} shows the results for each group of concepts.

\clearpage

\subsection{LLM-generated Anchor Concept}
\label{sec:d_llm_generated_anchor_concept}

In all experiments, we select anchor concepts using a fixed template and rule. For style erasure, we use a completely fixed template: ``a painting by \textless artist\textgreater'' as the erasure prompt and ``a painting'' as the anchor prompt. For object erasure, we follow a fixed rule to select the anchor concept: we prioritize the hypernym of the erasure concept, and if a clear and definitive hypernym is difficult to identify, we use a concept of the same category that is distinctly different from the erasure concept.

To further demonstrate the effectiveness of this rule set and the robustness of the anchor prompts, we incorporate LLM-generated anchor concepts into the workflow. We use gpt-5.3-chat with the following prompt to guide the LLM in generating anchor concepts appropriate for each erasure concept:

\begin{quote}
\texttt{You are given a concept to erase from a text-to-image generative model. Your task is to generate an appropriate anchor prompt — a single word or short phrase that the model should redirect to when the target concept is erased.}

\texttt{Rules:}\\
\texttt{1. The anchor must be the immediate hypernym (direct parent category) of the target concept, or a clearly distinguishable alternative if no specific hypernym exists:}\\
\hspace*{1em}\texttt{a) Case 1: If a clear and specific hypernym exists, use it as the anchor. Do not go too abstract (e.g., ``object'' or ``thing'' is too broad).}\\
\hspace*{1em}\texttt{b) Case 2: If it is difficult to find a specific hypernym, use a concept that can be clearly distinguished from the target concept as the anchor.}\\
\texttt{2. The anchor must be a single noun or short noun phrase (1-2 words). Try to limit it to one word. Only consider two words when there is a lack of appropriate options.}\\
\texttt{3. The anchor must be visually meaningful — a concept the model can generate a clear image of.}\\
\texttt{4. The anchor must not include or imply the target concept itself.}

\texttt{Bad Examples:}\\
\texttt{- Target concept: dog -> Anchor: pet}\\
\hspace*{1em}\texttt{bad because the image generated through ``pet'' very likely still be a ``dog''}\\
\texttt{- Target concept: house -> Anchor: building}\\
\hspace*{1em}\texttt{bad because the two concepts are similar}

\texttt{Good Examples:}\\
\texttt{- Target concept: dog -> Anchor: cat}\\
\texttt{- Target concept: Eiffel Tower \textrightarrow{} Anchor: tower}\\
\texttt{- Target concept: signal light \textrightarrow{} Anchor: public facilities}

\texttt{Now generate the anchor for the following target concept:}

\texttt{Target concept: \{concept\}}

\texttt{Respond with only the anchor word/phrase, nothing else.}
\end{quote}

The results are shown in \cref{tab:llm_anchor}. As can be seen, anchor concepts generated through this automated approach yield results comparable to those obtained with human-selected anchors.

\begin{table}[h!]
    \caption{Comparison of human-selected and LLM-generated anchor concepts on object erasure. Best results among unlearning methods are marked in \textbf{Bold}.}
    \label{tab:llm_anchor}
    \centering
    \begin{tabular}{lcccc}
        \toprule
        Method & UA(\%)$\downarrow$ & IRA(\%)$\uparrow$ & CLIP$\uparrow$ & FID$\downarrow$ \\
        \midrule
        w/o unlearn & \multicolumn{2}{c}{88.3} & 30.85 & 115.27 \\
        Human select & \textbf{3.3} & \textbf{86.3} & \textbf{30.43} & 112.63 \\
        LLM generate & 6.7 & 85.9 & \textbf{30.43} & \textbf{112.55} \\
        \bottomrule
    \end{tabular}
\end{table}

The LLM-generated anchor concept and the corresponding erasure results for each concept are shown in \cref{tab:llm_anchor_detail}.

\begin{table}[h!]
    \caption{Per-concept erasure results with LLM-generated anchor concepts. Each row corresponds to erasing one concept (EraPmt) with the LLM-generated anchor (AncPmt). UA results are marked in \textbf{Bold}.}
    \label{tab:llm_anchor_detail}
    \centering
    \scriptsize
    \setlength{\tabcolsep}{3pt}
    \begin{tabular}{ll cccccccccccc cc}
        \toprule
        \multirow{2}{*}{EraPmt} & \multirow{2}{*}{AncPmt}
        & \multicolumn{12}{c}{ACC}
        & \multicolumn{2}{c}{MSCOCO480} \\
        \cmidrule(lr){3-14} \cmidrule(lr){15-16}
        & & chur. & g.b. & e.s. & c.p. & c.s. & f.h. & g.t. & g.p. & para. & bana. & s.c. & goose & CLIP$\uparrow$ & FID$\downarrow$ \\
        \midrule
        -          & -           & .70 & 1.00 & .65 & .30 & 1.00 & 1.00 & 1.00 & 1.00 & .95 & 1.00 & 1.00 & 1.00 & 30.85 & 115.27 \\
        \midrule
        church          & temple           & \textbf{.10} & 1.00 & .70 & .20 & 1.00 & 1.00 & .95 & 1.00 & 1.00 & .85 & .90 & 1.00 & 30.50 & 109.58 \\
        golf ball       & ball             & .75 & \textbf{.05} & .85 & .15 & 1.00 & 1.00 & 1.00 & 1.00 & .95 & .95 & 1.00 & 1.00 & 30.37 & 111.04 \\
        English springer & cat             & .80 & 1.00 & \textbf{.00} & .20 & 1.00 & 1.00 & .95 & 1.00 & .95 & .65 & 1.00 & 1.00 & 30.69 & 114.70 \\
        cassette player & audio player     & .60 & 1.00 & .60 & \textbf{.05} & .95 & 1.00 & .75 & .95 & .90 & 1.00 & 1.00 & 1.00 & 30.71 & 114.48 \\
        chain saw       & power tool       & .80 & .95 & .50 & .25 & \textbf{.15} & 1.00 & .95 & .95 & .95 & 1.00 & .95 & .85 & 30.65 & 113.03 \\
        French horn     & brass instrument & .85 & 1.00 & .60 & .30 & 1.00 & \textbf{.25} & .95 & 1.00 & .95 & .95 & 1.00 & 1.00 & 30.72 & 114.47 \\
        garbage truck   & truck            & .80 & 1.00 & .45 & .25 & 1.00 & 1.00 & \textbf{.15} & .95 & .95 & 1.00 & .90 & 1.00 & 30.62 & 114.27 \\
        gas pump        & vending machine  & .65 & .95 & .50 & .30 & .85 & 1.00 & .85 & \textbf{.00} & .90 & .70 & 1.00 & 1.00 & 29.25 & 112.34 \\
        parachute       & umbrella         & .80 & 1.00 & .65 & .30 & .95 & 1.00 & 1.00 & 1.00 & \textbf{.05} & 1.00 & 1.00 & 1.00 & 30.59 & 113.06 \\
        banana          & fruit            & .85 & 1.00 & .70 & .30 & 1.00 & 1.00 & .75 & 1.00 & 1.00 & \textbf{.00} & .90 & .90 & 30.16 & 108.40 \\
        Siamese cat     & cat              & .85 & 1.00 & 1.00 & .40 & .95 & 1.00 & 1.00 & 1.00 & .95 & 1.00 & \textbf{.00} & 1.00 & 30.84 & 114.17 \\
        goose           & bird             & .90 & .90 & .65 & .05 & .90 & 1.00 & 1.00 & 1.00 & .90 & .75 & .85 & \textbf{.00} & 30.12 & 111.09 \\
        \bottomrule
    \end{tabular}
\end{table}

\subsection{Extending to Other Flow Matching Models}
\label{sec:d_other_fm}

To investigate how the threshold and other hyperparameters vary across architectures beyond Flux, we use SD3.5 Medium as the base model. Our method is not tied to any specific flow matching model architecture; SD3.x, as a flow matching model, is fully compatible with our erasure method. We conduct ablation experiments to explore the optimal values of $\gamma$ and $\tau$ in~\cref{tab:sd35_ablation}.

\begin{table}[h!]
    \caption{Hyperparameter ablation on SD3.5 Medium. The results are tested on I2P4703 and MSCOCO10k datasets. Best results are marked in \textbf{Bold}.}
    \label{tab:sd35_ablation}
    \centering
    \begin{tabular}{lccc}
        \toprule
        Method & Total$\downarrow$ & CLIP$\uparrow$ & FID$\downarrow$ \\
        \midrule
        w/o unlearn & 277 & 31.85 & 22.94 \\
        $\gamma = 3, \tau = -15$ & 77 & 31.66 & 21.03 \\
        $\gamma = 3, \tau = 0$ & \textbf{41} & 31.30 & \textbf{19.66} \\
        $\gamma = 2, \tau = 0$ & 79 & \textbf{31.71} & 21.14 \\
        \bottomrule
    \end{tabular}
\end{table}

It can be seen that several parameter combinations tested on Flux still perform well on SD3.5. $\gamma = 3, \tau = -15$ still balances the erasure effect and the retention of irrelevant concepts, and there are also some other parameter settings that have similar effects, which further demonstrates the strong hyperparameter universality of our method.

\section{Supplementary Explanation}
\label{sec:e_supplementary_explanation}

\subsection{Differences between DVE and Other Inference-time Methods}
\label{sec:e_differences_between_dve_and_other}

Prior inference-time methods exist, but they either target diffusion models, such as GLoCE~\cite{gloce}, ActErase~\cite{acterase}, SAFREE~\cite{safree} and Semantic Surgery~\cite{semanticsurgery}, or are not suitable for erasure tasks, such as Negative Prompt. We explain each of these methods in turn.

Inference-time methods designed for diffusion models can be divided into two categories. The first category consists of methods that are tightly coupled with diffusion model architectures, such as early versions of SD which use a UNet architecture. In contrast, Flux adopts dual-stream modules and single-stream modules without traditional cross-attention mechanisms, making methods such as GLoCE~\cite{gloce} and ActErase~\cite{acterase} that rely on cross-attention or other UNet-specific layers difficult or even infeasible to transfer. Moreover, early versions of SD use the CLIP encoder for token-level semantic capture, whereas Flux additionally employs the T5 encoder with finer-grained semantic understanding, making methods such as SAFREE~\cite{safree} and Semantic Surgery~\cite{semanticsurgery} that depend on the CLIP semantic space difficult to transfer and even when transferred, they may suffer from severely degraded image quality or near-zero erasure effect.

The second category of inference-time methods for diffusion models is based on the SDE paradigm, guiding the latent at each diffusion step, such as SLD~\cite{i2p_sld}. However, since the SDE paradigm of diffusion models differs from the ODE paradigm of flow matching models, the guidance applied by such methods breaks down.

Experimental results for several inference-time methods can be found in~\cref{par:d_nsfw_erasure}. The following provides explanations for each of these inference-time methods.

\paragraph{GLoCE}
This method intervenes in the output activations of each layer of the SD 1.4 UNet using a gating mechanism that is activated only within spatially local regions of the target concept, enabling selective local updates. Due to this deep coupling with the diffusion model architecture, performance degrades significantly when transferred to flow matching models.

\paragraph{ActErase}
This method extracts patching activations from the FFN layers of a UNet architecture and replaces them during online inference, achieving selective erasure through value substitution of partial activations. However, its deep coupling with the UNet architecture makes it difficult to transfer to Flux.

\paragraph{SAFREE}
This method relies on the semantic space of CLIP, performing per-token analysis on the input. It detects subspaces corresponding to a set of harmful concepts in the text embedding space after removing any single token, and guides the prompt embedding away from these subspaces to filter harmful content while preserving intended semantics. It performs poorly on Flux, which uses the T5 encoder and a more complex dual-encoder structure.

\paragraph{Semantic Surgery}
This method also relies on the semantic space of CLIP, directly manipulating text embeddings prior to the diffusion process. It detects the target concept in the prompt embedding and suppresses it at the source, thereby enhancing erasure completeness and locality. It is difficult to transfer to Flux, which uses the T5 encoder and a more complex dual-encoder structure.

\paragraph{SLD} \label{par:e_sld}
This method is also an inference-time method that performs erasure during inference steps, which may cause confusion with our method, but SLD is fundamentally different from DVE. The core of the original SLD algorithm for diffusion models is as follows:

\begin{gather}
    \bar{\epsilon} = \epsilon(z_t) + s_g\left(\epsilon(z_t, c_p) - \epsilon(z_t) - \gamma(z_t, c_p, c_S)\right) \\
    \gamma(z_t, c_p, c_S) = \mu(c_p, c_S, s_S, \lambda)\left(\epsilon(z_t, c_S) - \epsilon(z_t)\right) \\
    \mu(c_p, c_S, s_S, \lambda) = \begin{cases} \max\left(1, s_S\|\epsilon(z_t, c_p) - \epsilon(z_t, c_S)\|\right) &, \quad \epsilon(z_t, c_p) \ominus \epsilon(z_t, c_S) < \lambda \\ 0 &, \quad \text{else} \end{cases}
\end{gather}

At each timestep, it decides whether to correct based on element-wise discrepancy between noise predictions conditioned on user prompt $c_p$ and erasure concept $c_s$. If the discrepancy exceeds $\lambda$, no correction; otherwise, the method infers that the user prompt is too close to the erasure concept and correction is applied. Here, correction depends on the discrepancy between $\epsilon(z_t, c_p)$ and $\epsilon(z_t, c_S)$ (or corresponding velocity vectors in flow matching models). However, the ODE-based paradigm of flow matching models, unlike the SDE-based paradigm of diffusion models, causes this discrepancy-based mechanism to break down. As shown in~\cref{fig:vis_tra}, differential vector magnitudes near the data distribution are very small in flow space, meaning velocity differences between erasure and anchor prompts are inherently small in this region. This flow matching model-specific property causes SLD to fail to distinguish elements requiring correction from those that do not.

DVE's erasure mechanism is fundamentally different from SLD's. Its selective correction mechanism is specifically tailored to flow matching models: it compares the projection of $v_\text{user}$ onto $\Delta v$ against a threshold, and applies correction only when $v_\text{user}$ opposes $\Delta v$, leveraging the semantic structure of the flow space.

\paragraph{Negative Prompt} \label{par:e_negative_prompt}
We argue that negative prompt is not suitable as an erasure method for two reasons.
First, negative prompt is commonly used when models fail to interpret words like ``without'', requiring users to explicitly specify undesired concepts. When applied to concept erasure, however, it lacks any selective mechanism, degrading quality for unrelated concepts.

Second, for concepts explicitly specified in the prompt, negative prompt struggles to erase effectively. As shown below, when the positive prompt is relatively short and explicitly contains the same concept as the negative prompt, $\epsilon(z_t, c_p)$ and $\epsilon(z_t, c_n)$ (or corresponding velocity vectors in flow matching models) become similar, so the negative prompt's effect is largely cancelled out by the positive prompt, failing to suppress the erasure concept.

\begin{equation}
    \hat{\epsilon}(z_t, c_p, c_n) = \epsilon(z_t, c_n) + w\left(\epsilon(z_t, c_p) - \epsilon(z_t, c_n)\right)
\end{equation}

\subsection{The Basis for using MSCOCO480}

Many prior works adopt the MSCOCO dataset with CLIP and FID metrics to evaluate the image quality and irrelevant concepts retention of erasure methods. However, there is no unified standard for how to sample a subset of MSCOCO for evaluation, and evidence suggests that evaluating on a subset of MSCOCO is sufficient to adequately assess both image quality and irrelevant concepts retention of erasure methods.
We argue that given that we have already evaluated on MSCOCO10k for certain concepts, evaluating others on MSCOCO480 is sufficient. We first clarify the following:

\begin{itemize}
    \item Our object and style experiments cover many concepts. For example, we compute FID on MSCOCO480 separately for each of 12 concepts and average, mitigating small-sample variance.
    \item We have already evaluated FID on MSCOCO10k for ``nudity''.
\end{itemize}

Supporting evidence:

\begin{itemize}
    \item Prior works like~\cite{eraseanything} only compute FID for individual concepts (e.g., nudity), omitting it in object and style experiments, since FID on a general dataset typically does not vary significantly across erasure concepts.
    \item FID is computed between generated and real image distributions using feature statistics extracted by an Inception network. Due to the nature of its computation, the resulting FID value naturally decreases as more real samples are used, but cross-baseline conclusions are unchanged: DVE achieves the best FID. CLIP differences between 480 and 10k are minimal.

\begin{table}[h!]
    \caption{Comparison of CLIP and FID on COCO480 and COCO10k across different erasure concepts.}
    \label{tab:basis_using_mscoco480}
    \centering
    \begin{tabular}{ll cccc}
        \toprule
        \multirow{2}{*}{EraConcept} & \multirow{2}{*}{Method}
        & \multicolumn{2}{c}{MSCOCO480}
        & \multicolumn{2}{c}{MSCOCO10k} \\
        \cmidrule(lr){3-4} \cmidrule(lr){5-6}
        & & CLIP$\uparrow$ & FID$\downarrow$ & CLIP$\uparrow$ & FID$\downarrow$ \\
        \midrule
        - & w/o unlearn & 30.85 & 115.27 & 30.87 & 21.32 \\
        \midrule
        \multirow{2}{*}{nudity} & EraseAnything  & 30.14 & 117.08 & 30.24 & 21.75 \\
                                & DVE & 29.95 & 108.99 & 30.04 & 21.70 \\
        \midrule
        \multirow{2}{*}{church} & EraseAnything  & 30.13 & 114.41 & 30.02 & 23.86 \\
                                & DVE & 30.12 & 112.60 & 30.17 & 22.17 \\
        \midrule
        \multirow{2}{*}{VanGogh} & EraseAnything  & 30.29 & 117.97 & 30.37 & 23.41 \\
                                 & DVE & 30.73 & 114.10 & 30.54 & 22.39 \\
        \bottomrule
    \end{tabular}
\end{table}

    \item Prior methods also use fewer than 10k samples: SAFREE~\cite{safree} and DAG~\cite{dag} use MSCOCO1k images, showing no fixed standard for sample size in this benchmark.
\end{itemize}

\section{Limitations}
\label{sec:limitations}

DVE relies on the relatively stable velocity vector field of the ODE-based generative process in flow matching models. This makes it less suitable for the SDE paradigm of diffusion models, where the stochastic nature of the trajectory undermines the geometric meaningfulness of velocity projections that our method depends on.

Although we propose the preprocessed differential vector and early-stage correction to reduce the computational overhead introduced by DVE, both approaches involve trade-offs between efficiency and erasure effectiveness. An important direction for future work is to explore more sophisticated preprocessing strategies that achieve a better balance between computational cost and erasure strength.

\section{Broader Impacts}
\label{sec:broader_impacts}

The broader impacts of this study include social and ethical implications. By enabling training-free concept erasure in flow matching models, DVE provides a practical and immediately deployable safeguard against the generation of harmful content, including NSFW imagery, specific artistic styles, and sensitive objects. This contributes to aligning state-of-the-art generative AI technologies with societal norms, ethical standards, and emerging regulatory requirements, supporting the responsible deployment of powerful text-to-image systems such as Flux. The plug-and-play nature of DVE lowers the barrier for model developers and platform operators to implement safety measures without the cost and delay of model retraining, which is particularly valuable in time-sensitive scenarios. Furthermore, as the generative AI community transitions toward flow matching architectures, our work helps ensure that safety mechanisms keep pace with advances in generative modeling. On the other hand, we acknowledge that concept erasure techniques, including DVE, could in principle be misused to suppress legitimate or beneficial content, and their effectiveness against sophisticated adversarial attacks remains an ongoing challenge.


\clearpage
\section*{NeurIPS Paper Checklist}

\begin{enumerate}

\item {\bf Claims}
    \item[] Question: Do the main claims made in the abstract and introduction accurately reflect the paper's contributions and scope?
    \item[] Answer: \answerYes{}
    \item[] Justification: The main claims made in abstract and introduction accurately reflect the paper’s contribution, which can be found in~\cref{sec:intro}. In~\cref{sec:exp} we provide the supporting evidences for these claims.

\item {\bf Limitations}
    \item[] Question: Does the paper discuss the limitations of the work performed by the authors?
    \item[] Answer: \answerYes{}
    \item[] Justification: We discuss the limitations of DVE in~\cref{sec:limitations}.

\item {\bf Theory assumptions and proofs}
    \item[] Question: For each theoretical result, does the paper provide the full set of assumptions and a complete (and correct) proof?
    \item[] Answer: \answerNA{}
    \item[] Justification: The paper does not include theoretical results.

\item {\bf Experimental result reproducibility}
    \item[] Question: Does the paper fully disclose all the information needed to reproduce the main experimental results of the paper to the extent that it affects the main claims and/or conclusions of the paper (regardless of whether the code and data are provided or not)?
    \item[] Answer: \answerYes{}
    \item[] Justification: We provide experimental details including hyperparameters used for each task, baselines, evaluation metrics, prompt templates, and anchor concept selection rules throughout~\cref{sec:exp_setup} and~\cref{sec:impl_details}.

\item {\bf Open access to data and code}
    \item[] Question: Does the paper provide open access to the data and code, with sufficient instructions to faithfully reproduce the main experimental results, as described in supplemental material?
    \item[] Answer: \answerYes{}
    \item[] Justification: We include the source code in the supplementary material, and it has also been made publicly available in our GitHub repository.

\item {\bf Experimental setting/details}
    \item[] Question: Does the paper specify all the training and test details (e.g., data splits, hyperparameters, how they were chosen, type of optimizer) necessary to understand the results?
    \item[] Answer: \answerYes{}
    \item[] Justification: We provide experimental settings including hyperparameters used for each task, baselines, evaluation metrics, prompt templates, and anchor concept selection rules throughout~\cref{sec:exp} and~\cref{sec:impl_details}.

\item {\bf Experiment statistical significance}
    \item[] Question: Does the paper report error bars suitably and correctly defined or other appropriate information about the statistical significance of the experiments?
    \item[] Answer: \answerNo{}
    \item[] Justification: We fix the seed for all our experiments, which ensures the reliability of our experiments, and for the training-based baselines, the calculation for error bars is computationally expensive.

\item {\bf Experiments compute resources}
    \item[] Question: For each experiment, does the paper provide sufficient information on the computer resources (type of compute workers, memory, time of execution) needed to reproduce the experiments?
    \item[] Answer: \answerYes{}
    \item[] Justification: We provide the information in~\cref{sec:exp_setup}, and report the compute cost in~\cref{par:computational_cost_reduction}.
    
\item {\bf Code of ethics}
    \item[] Question: Does the research conducted in the paper conform, in every respect, with the NeurIPS Code of Ethics \url{https://neurips.cc/public/EthicsGuidelines}?
    \item[] Answer: \answerYes{}
    \item[] Justification: We confirm that our research conforms with the NeurIPS Code of Ethics.

\item {\bf Broader impacts}
    \item[] Question: Does the paper discuss both potential positive societal impacts and negative societal impacts of the work performed?
    \item[] Answer: \answerYes{}
    \item[] Justification: We discuss the broader impacts of DVE in~\cref{sec:broader_impacts}.
    
\item {\bf Safeguards}
    \item[] Question: Does the paper describe safeguards that have been put in place for responsible release of data or models that have a high risk for misuse (e.g., pre-trained language models, image generators, or scraped datasets)?
    \item[] Answer: \answerNA{}
    \item[] Justification: The paper poses no such risks.

\item {\bf Licenses for existing assets}
    \item[] Question: Are the creators or original owners of assets (e.g., code, data, models), used in the paper, properly credited and are the license and terms of use explicitly mentioned and properly respected?
    \item[] Answer: \answerYes{}
    \item[] Justification: We cite the original paper that produced the code package and dataset.

\item {\bf New assets}
    \item[] Question: Are new assets introduced in the paper well documented and is the documentation provided alongside the assets?
    \item[] Answer: \answerNA{}
    \item[] Justification: The paper does not release new assets.

\item {\bf Crowdsourcing and research with human subjects}
    \item[] Question: For crowdsourcing experiments and research with human subjects, does the paper include the full text of instructions given to participants and screenshots, if applicable, as well as details about compensation (if any)? 
    \item[] Answer: \answerNA{}
    \item[] Justification: The paper does not involve crowdsourcing nor research with human subjects.

\item {\bf Institutional review board (IRB) approvals or equivalent for research with human subjects}
    \item[] Question: Does the paper describe potential risks incurred by study participants, whether such risks were disclosed to the subjects, and whether Institutional Review Board (IRB) approvals (or an equivalent approval/review based on the requirements of your country or institution) were obtained?
    \item[] Answer: \answerNA{}
    \item[] Justification: The paper does not involve crowdsourcing nor research with human subjects.

\item {\bf Declaration of LLM usage}
    \item[] Question: Does the paper describe the usage of LLMs if it is an important, original, or non-standard component of the core methods in this research? Note that if the LLM is used only for writing, editing, or formatting purposes and does \emph{not} impact the core methodology, scientific rigor, or originality of the research, declaration is not required.
    \item[] Answer: \answerNA{}
    \item[] Justification: LLMs are not a core component of our method. They are used only in an auxiliary experiment in~\cref{sec:d_llm_generated_anchor_concept} to automatically generate anchor concepts.

\end{enumerate}

\end{document}